\documentclass{article}

\usepackage[utf8]{inputenc} 
\usepackage[T1]{fontenc}    
\usepackage[preprint,nohyperref]{icml2026}
\usepackage{xurl}
\usepackage{hyperref}
\usepackage{url}            
\usepackage{booktabs}       
\usepackage{amsfonts}       
\usepackage{nicefrac}       
\usepackage{microtype}      
\usepackage{xcolor}         
\usepackage{graphicx}
\usepackage{subcaption}
\usepackage{amsmath}
\usepackage{amssymb}
\usepackage{mathtools}
\usepackage{amsthm}
\usepackage{colortbl}
\usepackage{enumitem}
\usepackage{pifont}
\usepackage{listings}
\usepackage{multirow}
\usepackage{textcomp}
\usepackage{wrapfig}
\usepackage{algorithm}
\usepackage{algorithmic}
\usepackage[skins,breakable]{tcolorbox}
\usepackage[capitalize,noabbrev]{cleveref}
\usepackage{float}                        
\usepackage{placeins}                      
\usepackage{needspace}

\definecolor{nvgreen}{RGB}{118, 185, 0}
\definecolor{techdark}{RGB}{30, 30, 30}
\definecolor{techlight}{RGB}{245, 245, 245}
\definecolor{codebg}{rgb}{0.95,0.95,0.95}

\newtcolorbox{chatbox}[2][]{
  enhanced,
  colback=techlight,
  colframe=gray!60!white,
  colbacktitle=gray!15!white,
  coltitle=black,
  title={#2},
  fonttitle=\sffamily\bfseries\scriptsize,
  fontupper=\sffamily\small,
  boxrule=0.5pt,
  arc=1mm,
  left=2mm, right=2mm, top=1mm, bottom=1mm,
  borderline west={2pt}{0pt}{nvgreen},
  #1
}

\newtcolorbox{promptbox}[1]{
  enhanced,
  breakable,
  colback=gray!3,
  colframe=gray!35,
  colbacktitle=gray!10,
  coltitle=black,
  title={#1},
  fonttitle=\small\bfseries,
  fontupper=\small,
  before upper={\setlength{\parindent}{0pt}\setlength{\parskip}{3pt}},
  boxrule=0.5pt,
  arc=1mm,
  left=2mm, right=2mm, top=1.5mm, bottom=1.5mm,
  borderline west={1.5pt}{0pt}{nvgreen},
  before skip=6pt,
  after skip=6pt,
}

\newtcolorbox{outerbox}[1][]{
  enhanced,
  colback=white,
  colframe=techdark,
  colbacktitle=nvgreen,
  coltitle=white,
  fonttitle=\sffamily\bfseries\large,
  sharp corners,
  boxrule=1.0pt,
  drop shadow southeast,
  breakable,
  #1
}

\theoremstyle{plain}

\theoremstyle{definition}

\theoremstyle{remark}

\definecolor{preprintlink}{rgb}{0,0.08,0.45}
\hypersetup{colorlinks=true,allcolors=preprintlink,
  pdfauthor={Guo Chen, Lidong Lu, Yicheng Liu, Liangrui Dong, Lidong Zou, Jixin Lv, Zhenquan Li, Xinyi Mao, Baoqi Pei, Shihao Wang, Zhiqi Li, Karan Sapra, Fuxiao Liu, Yin-Dong Zheng, Yifei Huang, Limin Wang, Zhiding Yu, Andrew Tao, Guilin Liu, Tong Lu},
  pdftitle={Towards Multimodal Lifelong Understanding: A Multi-Scale Benchmark}, pdfsubject={arXiv preprint}}


\icmltitlerunning{Towards Multimodal Lifelong Understanding: A Multi-Scale Benchmark}

\begin{document}

\makeatletter
\let\preprint@sict\@sict
\def\@sict#1#2#3#4#5#6[#7]#8{%
  \NR@gettitle{#7}%
  \preprint@sict{#1}{#2}{#3}{#4}{#5}{#6}[#7]{#8}}
\makeatother

\twocolumn[
  \icmltitle{Towards Multimodal Lifelong Understanding:\texorpdfstring{\\}{ }A Multi-Scale Benchmark}

  \begin{center}
    \vspace{-0.2in}

    \normalsize \bfseries
    Guo Chen$^{1,2*}$,
    Lidong Lu$^{1*}$,
    Yicheng Liu$^1$,
    Liangrui Dong$^1$,
    Lidong Zou$^1$,
    Jixin Lv$^1$,
    Zhenquan Li$^1$, \\
    Xinyi Mao$^1$,
    Baoqi Pei$^{3}$,
    Shihao Wang$^{2}$,
    Zhiqi Li$^{2}$,
    Karan Sapra$^{2}$,
    Fuxiao Liu$^{2\S}$,
    Yin-Dong Zheng$^{4\S}$, \\
    Yifei Huang$^{5\S}$,
    Limin Wang$^{1\S}$,
    Zhiding Yu$^{2\S}$,
    Andrew Tao$^{2\S}$,
    Guilin Liu$^{2\S}$,
    Tong Lu$^{1\S}$

    \vspace{0.5em}

    \small \mdseries
    $^1$Nanjing University \quad
    $^2$NVIDIA \quad
    $^3$Zhejiang University \\
    $^4$Shanghai Jiao Tong University \quad
    $^5$The University of Tokyo

    \vspace{0.6em}

    \normalsize 
    \href{https://huggingface.co/datasets/MM-Lifelong/MM-Lifelong}{\texttt{\textbf{[Dataset]}}}

  \end{center}

  \icmlkeywords{Machine Learning, ICML}
  \vskip 0.35em
  {\begin{minipage}{\textwidth}
    \centering
    \captionsetup{type=figure}
    \includegraphics[width=\linewidth]{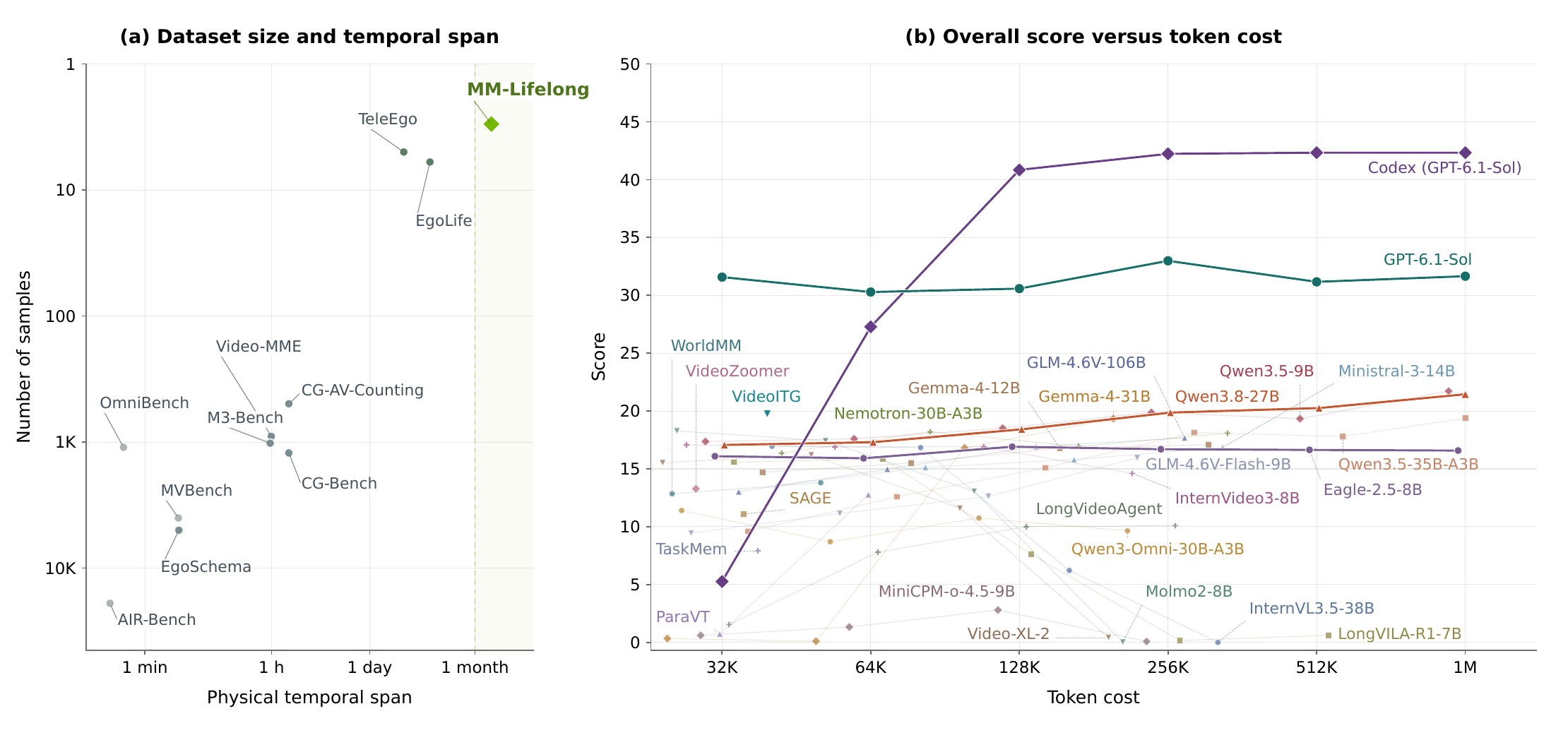}
    \caption{\textbf{Overview of MM-Lifelong.} \textbf{(a)} Dataset sample count and maximum physical temporal span; MM-Lifelong covers 51 days. \textbf{(b)} Overall answer score versus token cost up to 1M tokens. Table~\ref{tab:main_results} reports Score and Ref@300 for 16 MLLMs and 8 agents.}
    \label{fig:overall}
  \end{minipage}}
  \vskip 1em
  {\begin{abstract}
Multimodal systems can access increasingly long video histories, but their ability to connect observations of the same subject across days and months remains insufficiently evaluated. We formalize the \textbf{Lifelong Horizon}, distinguishing recorded duration from the physical temporal span of a persistent subject's history. To evaluate this setting, we introduce \textbf{MM-Lifelong}, a benchmark with 181.1 hours of video and 1289 questions across Day, Week, and Month scales. Its Month subset contains 105.6 hours of footage spanning 51 days. Every question is paired with manually annotated clue intervals. We evaluate answer quality and temporal grounding separately, using question-specific rubrics to accommodate valid alternative answers and partial correctness. Increasing the input budget yields uneven gains in answer quality for multimodal large language models (MLLMs). Among agents, answer quality and temporal grounding yield different rankings. Error analysis identifies lost answer-relevant details, confusion between event instances, and unresolved conflicting observations. These findings motivate memory systems that preserve event-specific evidence and verification that checks its support for the final answer.
\end{abstract}}
]

\icmlcorrespondingauthor{Tong Lu}{lutong@nju.edu.cn}

\printAffiliationsAndNotice{%
    \textsuperscript{*}Equal contribution.
    $^\S$Paper advisor%
}

\section{Introduction}
\label{sec:intro}

Recent advances in multimodal large language models (MLLMs) have expanded the amount of video that models can process within a single context~\cite{yang2025qwen3,gemini}. In parallel, agentic systems use retrieval, external memory, and tools to access observations beyond that context~\cite{zhang2025dvd,zuo2025videolucy}. These developments create an opportunity to study multimodal understanding over a subject's extended history. Across days and months, similar activities recur, attributes and locations change, and observations are interrupted by recording gaps. Answering questions about such histories requires identifying specific event instances and relating evidence across recording sessions. We study this capability as \textbf{multimodal lifelong understanding}.

Existing benchmarks provide only partial coverage of this setting. Long-video benchmarks such as Video-MME~\cite{videomme} and LongVideoBench~\cite{wu2024longvideobench} evaluate understanding within individual recordings. Multi-video benchmarks~\cite{peng2025mvu-eval,zhu2025cvbench} broaden the input to multiple clips, which need not follow a persistent subject along a shared timeline. EgoLife~\cite{yang2025egolife} and TeleEgo~\cite{yan2025teleego} establish longitudinal evaluation over several days, providing an important foundation for studying subject-specific histories. Month-scale histories extend this setting to recurring events and changing contexts distributed across substantially longer physical timelines. Evaluating them requires distinguishing \emph{observational duration}, the amount of recorded footage, from \emph{physical temporal span}, the elapsed time covered by those observations. We formalize this distinction through the \textbf{Lifelong Horizon}: sustained observation of a persistent subject across time.

To evaluate this setting, we introduce \textbf{MM-Lifelong}, a benchmark comprising 181.1 hours of video and 1289 questions across Day-scale gameplay, Week-scale egocentric life, and Month-scale livestreams. The Month subset contains 105.6 hours of footage distributed across 51 days, exposing a substantial separation between observed playback and physical temporal span (Figure~\ref{fig:overall}(a)). Questions probe both the recovery of specific events from extensive histories and the integration of observations from disjoint intervals. Every question is paired with manually annotated clue intervals that identify its supporting evidence. We evaluate answer quality and temporal localization separately, using question-specific rubrics to accommodate valid alternative answers and partial correctness. This design supports examining both what a system answers and where it can locate the relevant evidence.

We evaluate end-to-end MLLMs and agentic systems without benchmark-specific training. The best evaluated MLLM scores 32.98 out of 100 at 256K tokens, and increasing input budgets produces uneven gains across models (Figure~\ref{fig:overall}(b)). Among agents, stronger temporal grounding does not consistently accompany better answers. Analysis of representative failures identifies three recurring problems: loss of answer-relevant details, confusion between event instances, and failure to resolve conflicting observations. These findings identify concrete requirements for lifelong multimodal systems: preserving event-specific evidence, maintaining its temporal associations, and verifying the observations supporting an answer.

\section{Related Work}

\textbf{Multimodal Understanding Benchmarks.} Multimodal evaluation has progressed from static single-image tasks (e.g., MMMU~\cite{yue2024mmmu}, MMBench~\cite{liu2024mmbench}) to dynamic video understanding. Early video benchmarks focused on short-term recognition~\cite{mvbench,ning2023videobench}, while more recent works such as VideoMME~\cite{videomme} and LongVideoBench~\cite{wu2024longvideobench} reach hour-long durations. Multi-video benchmarks~\cite{peng2025mvu-eval,zhu2025cvbench} extend the horizon further but typically aggregate disjointed clips that share no continuous timeline. Building on pioneering egocentric datasets~\cite{ego4d,egoexolearn,he2025egoexobench,pei2025egothinker}, EgoLife~\cite{yang2025egolife} pushes into longitudinal first-person recording but stays within a single shared apartment. MM-Lifelong complements these efforts with 105.6 hours of continuous live broadcasts spanning 51 days, in which physical timeline far exceeds observed playback.

\textbf{Benchmarks for Long-Context Memory.} In text, benchmarks like LongBench~\cite{bai2024longbench} and BABILong~\cite{kuratov2024babilong} use massive contexts for state tracking but lack visual input. Multimodal counterparts either rely on discrete images (e.g., Mem-Gallery~\cite{bei2026mem-gallery}) or focus on short-term streaming responsiveness~\cite{yang2025svbench}, leaving the long-horizon, continuously evolving multimodal setting largely untested. MM-Lifelong fills this gap with ``Needle-in-a-Lifestream'' and multi-hop tasks that probe whether models can track a single subject's state across weeks of unobserved gaps.

\textbf{Working and Persistent Memory.} Two complementary lines of work address the burden of long-context video. \emph{Within the model}, the field is moving from KV-cache compression and eviction~\cite{chen2025eagle,xiao2023streamingllm} to architectural changes that decouple memory footprint from sequence length, such as linear and hybrid attention~\cite{yang2025qwen3,blakeman2025nemotron-h} and latent or conditional memory~\cite{liu2024deepseekv3,cheng2026deepseek-engram}. \emph{Outside the model}, agentic systems transcend finite context via recursive tool use and external memory orchestration~\cite{gemini,google_2025_deepsearch,chhikara2025mem0}, supported by retrieval mechanisms~\cite{faysse2024colpali,wan2025mmgraphrag,rege2026agentic} and video-specific agents that perform frame selection and persistent visual memory~\cite{wang2025videoitg,chen2025lvagent,jin2025videomem,chen2026telemembuildinglongtermmultimodal,wang2025omemomnimemorypersonalized,chen2026iterresearchrethinkinglonghorizonagents,yu2026agenticmemorylearningunified,huang2025vinci}. Existing benchmarks remain largely text-centric or limited to discrete visual tasks. MM-Lifelong evaluates both directions on the same continuous, 100+ hour stream, characterizing context use within MLLMs and evidence access within agentic systems.

\section{Multimodal Lifelong Understanding}
\label{sec:problem_formulation}

Multimodal lifelong understanding extends long-video understanding to streams whose physical timeline far exceeds what the model can actually observe, so that the model must reason across stretches of real time it never saw. This section makes this precise: we define the relevant temporal quantities and the conditions under which a task qualifies as lifelong rather than merely long.

\begin{figure}[t]
    \centering
    \includegraphics[width=\linewidth]{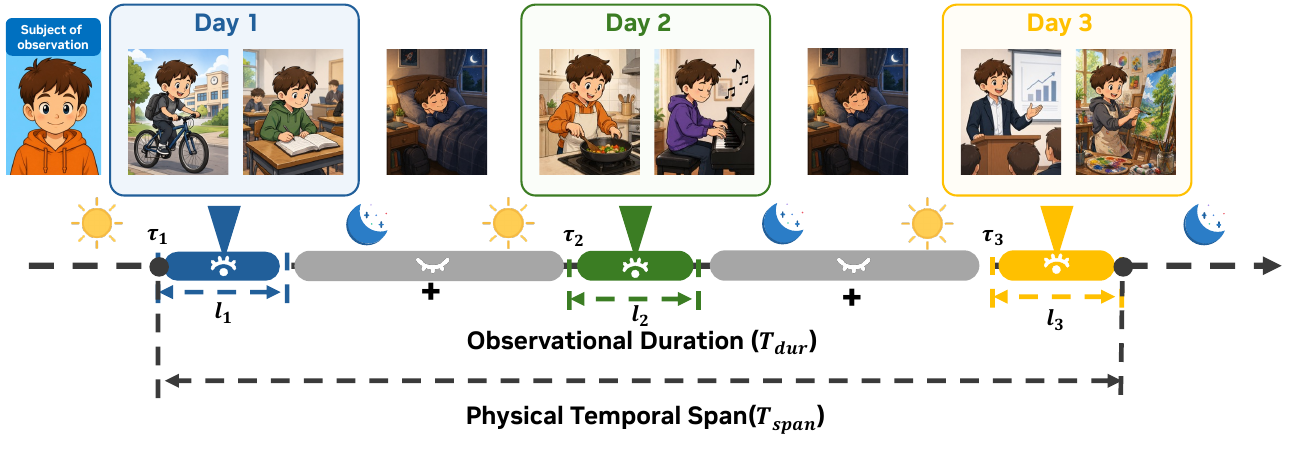}
    \caption{\textbf{Problem formalization of Multimodal Lifelong Understanding.} A persistent subject is observed across multiple recording sessions on different days, with sleep cycles and off-camera periods between sessions leaving unobserved gaps. Each recorded clip has a playback length $l_i$ and a real-world start time $\tau_i$. Two temporal quantities characterize the dataset: \textit{Observational Duration} $T_{dur}=\sum_i l_i$ (total playback, summed across clips) and \textit{Physical Temporal Span} $T_{span}=(\tau_N+l_N)-\tau_1$ (chronological horizon from first to last clip). On lifelong timelines spanning days to months, continuous capture is infeasible, so $T_{span}\!\gg\!T_{dur}$ arises naturally and forces models to reason across large unobserved gaps.}
    \label{fig:task_def}
\end{figure}

A multimodal dataset is a discrete sampling of a continuous physical timeline, consisting of $N$ video clips. Each clip has a playback length $l_i$ and a real-world starting timestamp $\tau_i$. The pair $(l_i, \tau_i)$ separates two temporal quantities that benchmarks typically conflate: how much footage the model sees, and what real-world span that footage covers.

We characterize the temporal scale of a dataset by two metrics (Figure~\ref{fig:task_def}):
\begin{itemize}[nosep]
    \item \textbf{Observational Duration ($T_{dur}$):} The sum of the playback lengths of all observed clips: $T_{dur} = \sum_{i=1}^{N} l_i$.
    \item \textbf{Physical Temporal Span ($T_{span}$):} The chronological horizon from the start of the first clip to the end of the last: $T_{span} = (\tau_N + l_N) - \tau_1$.
\end{itemize}
Long-video benchmarks typically span hours, allowing clips to be packed densely so that $T_{span} \approx T_{dur}$. Lifelong datasets, however, span days, weeks, or months: at this scale continuous capture is no longer feasible, since sleep cycles, off-camera periods, and the cost of round-the-clock recording inevitably create gaps. As a result, $T_{span} \gg T_{dur}$ arises naturally on lifelong streams, with the unobserved gaps representing real-world time that simply cannot be recorded. The model must therefore infer what happened during stretches it never saw, rather than locate adjacent frames it did see.

$T_{span} \gg T_{dur}$ alone is not sufficient: a five-minute clip from a year-long timeline already satisfies $T_{span}/T_{dur} = 10^5$, yet is just short-form recognition with arbitrary timestamps. Three additional constraints on the dataset rule out such constructions:
\begin{enumerate}[nosep, leftmargin=*]
    \item \textbf{Daily-scale observation ($T_{dur} \ge 12h$):} each recording covers a full diurnal routine (roughly a person's active period), not isolated fragments.
    \item \textbf{Cross-day temporal span ($T_{span} \ge 24h$):} the span includes at least one unobserved interval such as a sleep cycle or off-camera period. A 23-hour recording does not qualify; a 26-hour span with a sleep cycle does.
    \item \textbf{Persistent subject:} a single identity is tracked continuously across $T_{span}$; otherwise the task collapses to aggregating across unrelated subjects. Appearance, location, and activity may change, but the agent does not.
\end{enumerate}
A task satisfying all three lies in the \textit{Lifelong Horizon}.

\begin{table*}[t]
\centering
\footnotesize
\caption{\textbf{Multi-scale design of MM-Lifelong.} Each scale tracks one subject's evolving state: a gamer, wearer, or streamer. At the month scale, $T_{span}$ exceeds that of released video corpora by 2--3 orders of magnitude.}
\label{tab:continuum_levels}
\setlength{\tabcolsep}{4pt}
\begin{tabular*}{\textwidth}{@{\extracolsep{\fill}}lllccp{5.4cm}@{}}
\toprule
\textbf{Scale} & \textbf{Domain} & \textbf{Subject} & \textbf{$T_{dur}$} & \textbf{$T_{span}$} & \textbf{Data Source Description} \\
\midrule
\textit{Day} & Gamer's Journey & Protagonist & 23.6h & $\sim$24h & Complete narrative walkthrough tracking the avatar's inventory and skill progression. \\
\textit{Week} & Egocentric Life & Wearer & 51.9h & $\sim$7d & Continuous first-person recording of daily routines and household interactions from EgoLife~\cite{yang2025egolife}. \\
\textit{Month} & Live Stream & Streamer & 105.6h & $\sim$51d & Unscripted IRL stream tracking the influencer's travel across cities and social events. \\
\bottomrule
\end{tabular*}
\end{table*}

Continuous multimodal footage at this scale is rare in practice. The most ambitious longitudinal efforts (EgoLife $\sim$7 days, TeleEgo $\sim$3 days) reach the day scale at most, leaving the month-scale Lifelong Horizon largely unpopulated. Section~\ref{sec:dataset} introduces the recordings and annotation pipeline through which we populate this horizon at three scales (Day, Week, Month).

\section{The MM-Lifelong Benchmark}
\label{sec:dataset}

\begin{figure*}[t]
    \centering
    \vspace{-0.5em}
    \includegraphics[width=\textwidth]{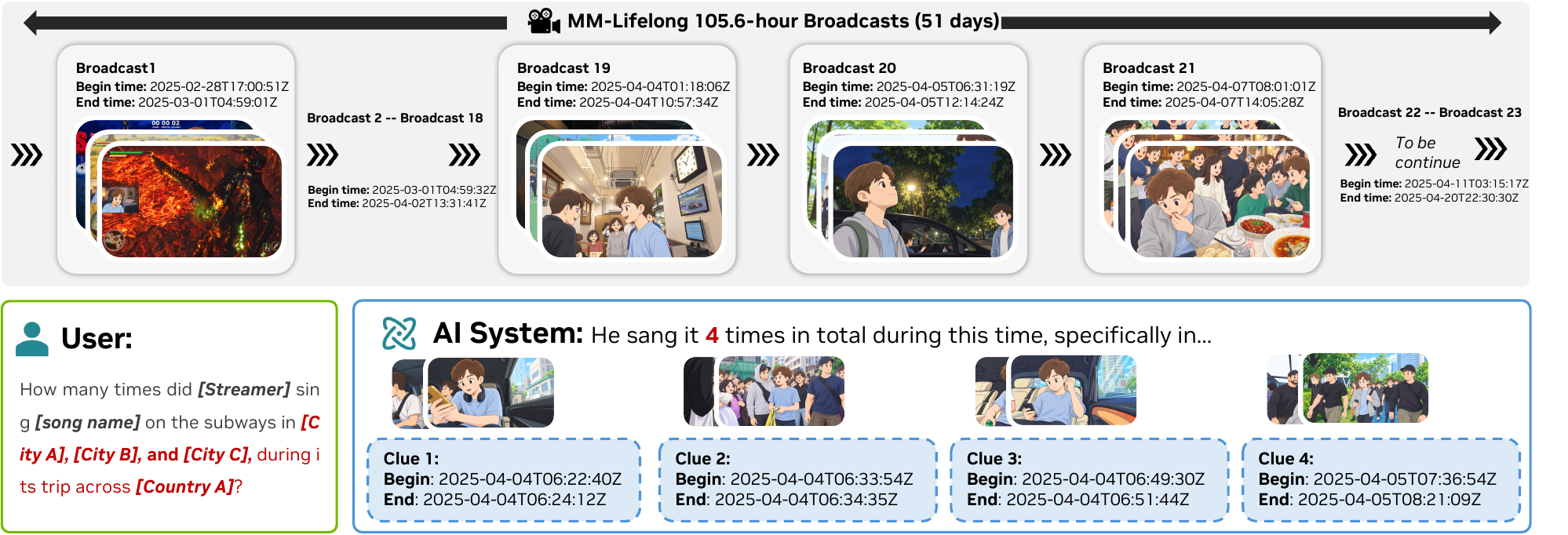}
    \vspace{-0.5em}
    \caption{\textbf{(a)} The Live Stream subset of \textbf{MM-Lifelong} comprises 105.6 hours of broadcast footage spanning 51 days. \textbf{(b)} A multi-clue (multi-hop) reasoning question with a long temporal certificate: the task requires identifying all occurrences where the streamer sings a specific song on subways across multiple cities. Answering correctly requires persistent memory and multi-event inference across more than 10 hours of livestream data. People are depicted as fictional characters.}
    \label{fig:teaser}
\end{figure*}

We introduce \textbf{MM-Lifelong}, a \textit{Multi-Scale Benchmark} that operationalizes the lifelong setting defined in Section~\ref{sec:problem_formulation}. Since continuous lifetime recording is infeasible at scale, MM-Lifelong approximates this setting through three complementary scales (Day, Week, Month) that together cover the temporal properties of the Lifelong Horizon.

\subsection{Dataset Construction}
\label{subsec:construction}

We describe the multi-scale design, video collection, and clue-grounded annotation of MM-Lifelong.

\textbf{Multi-scale design.}
Lifelong complexity requires content variation as well as duration. Our three recording domains span distinct ratios of Observational Duration ($T_{dur}$) to Physical Span ($T_{span}$; Table~\ref{tab:continuum_levels}). \textit{Day} and \textit{Week} provide dense observations with minimal interruption, while \textit{Month} captures sparse long-term evolution with large unobserved gaps ($T_{span} \gg T_{dur}$). We exclude year-scale data because of contamination through frontier-model parametric knowledge; Appendix~\ref{app:dataset:data-contamination} provides the analysis and examples.

\textbf{Why depth over breadth.}
We prioritize temporal depth over multiple subjects per scale for three reasons. (i) Lifelong reasoning requires \emph{intra-subject} state accumulation: tracking appearance, knowledge, and routines across unobserved gaps. Shallow clips from many subjects test a different capability, covered by multi-video benchmarks~\cite{peng2025mvu-eval,zhu2025cvbench}. (ii) Whereas EgoLife~\cite{yang2025egolife} and TeleEgo~\cite{yan2025teleego} each cover one egocentric domain, our three scales follow different subjects in disjoint domains (gamer, wearer, streamer), preventing a single-persona shortcut across all three. (iii) At fixed annotation density, recruiting and clue-grounding multi-week streams from $N>1$ subjects requires roughly $N$ times the annotator-months. Existing cross-domain coverage and the benchmark's unsaturated performance favor temporal depth before this costly expansion.

\textbf{Video Diversity and Collection.}
The \textit{Gaming} (synthetic) and \textit{Egocentric} (first-person routine) subsets cover specialized scenarios. \textit{Live Stream} adds unscripted open-world content, from \textit{indoor chatting, gaming, and reaction videos} to \textit{outdoor vlogs, sports, chaotic events, and singing performances}. This diversity tests robustness across visual contexts as well as temporal scales. The raw collection totals 211 GB of video.

\subsection{Annotation Protocol}

To support reproducible evaluation and future extensions, we adopt a \textit{clue-grounded annotation strategy} inspired by CG-Bench~\cite{chen2024cg}.
Unlike traditional QA pairs that provide only the final answer, we explicitly annotate the \textit{clue intervals}, the specific video segments containing the visual evidence required for reasoning.
This grounded approach both facilitates automated evaluation (Section~\ref{sec:eval}) and leaves room for future interpretability studies that inspect which evidence a model attended to.

\textbf{Task Definition.} Building on the clue-grounded framework, we design two task families that probe complementary aspects of lifelong understanding.
\textbf{Type I: Needle-in-a-Lifestream} targets single-occurrence details embedded in long streams; models must identify unique, low-frequency events within 100+ hour streams (e.g., \textit{``exact moment the camera dropped''}), testing precise localization and robustness to surrounding noise.
\textbf{Type II: Multi-Hop Reasoning} requires aggregating information across disjoint intervals separated by hours or days (e.g., \textit{``outfit change between check-in and dinner''}), which requires maintaining a persistent state and performing logical inference rather than single-shot retrieval.

\textbf{Quality Control.} We apply a three-stage pipeline to ensure dataset integrity.
\textbf{1) Distribution Enforcement:} Instead of random sampling, we curate data to ensure diverse clue durations and validate \textit{temporal certificates} following EgoSchema~\cite{mangalam2023egoschema}, reserving a subset for \textit{ultra-long dependencies ($>10$h)} to force cross-session inference.
\textbf{2) Multi-Stage Verification:} All triplets undergo a two-stage validation: manual expert cross-checks eliminate ambiguity, and a GPT-5-based~\cite{openai2026gpt5} filter removes questions answerable by common sense, so that answers depend on the visual evidence rather than prior knowledge.
\textbf{3) Human-Human Agreement:} Three independent evaluators (not the original annotators) re-answer each question with full video access. Their per-question average is reported as the ``Human'' row in Table~\ref{tab:main_results} ($>80\%$ accuracy across all splits), confirming that the annotated answers are reproducible by independent humans.

\subsection{Dataset Statistics}
\label{subsec:stats}

\begin{figure}[t]
\centering
\begin{minipage}[b]{0.48\linewidth}
\centering
\includegraphics[width=\linewidth]{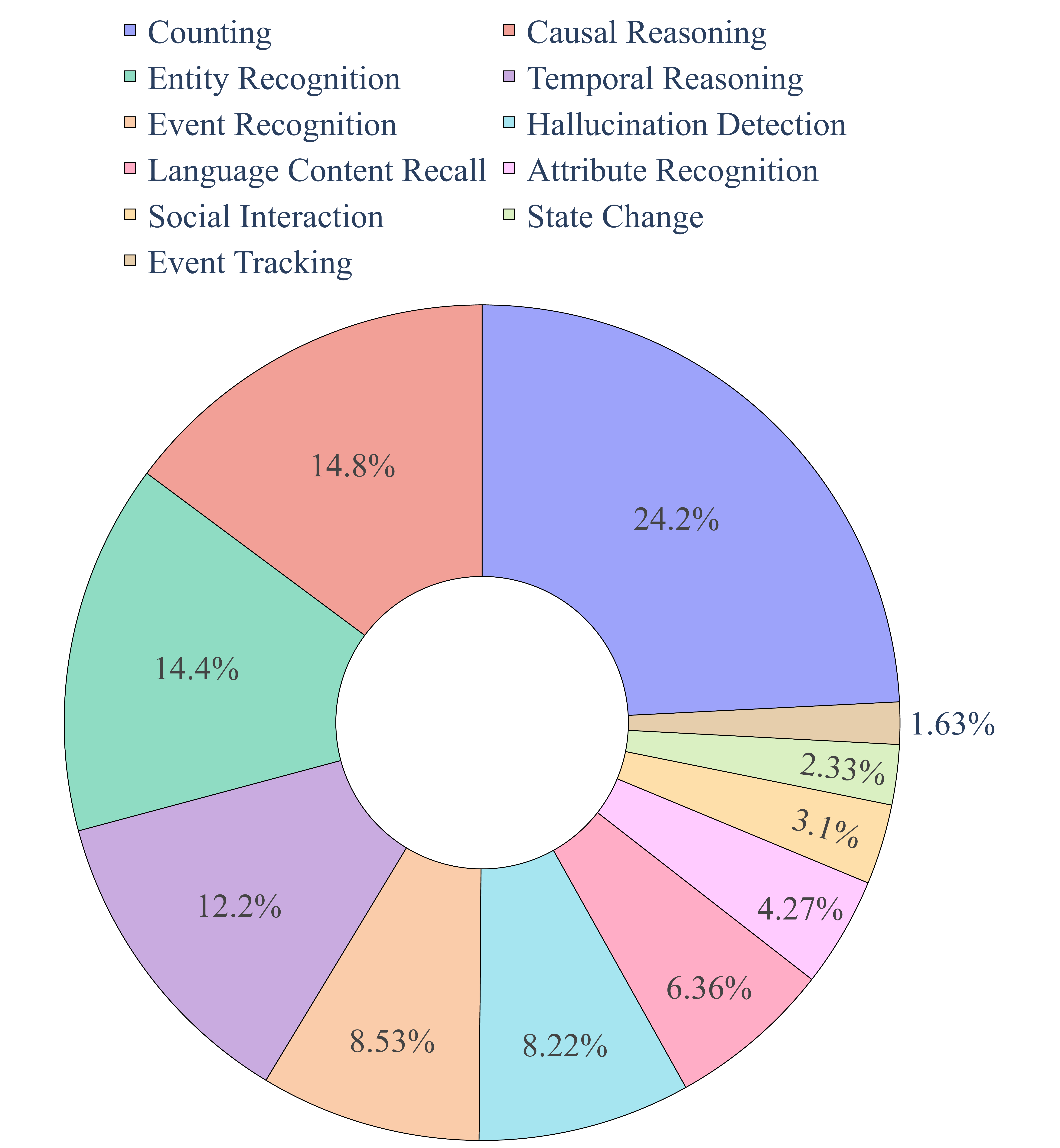}
\caption{Distribution of question categories.}
\label{fig:question-categroy}
\end{minipage}
\hfill
\begin{minipage}[b]{0.48\linewidth}
\centering
\includegraphics[width=\linewidth]{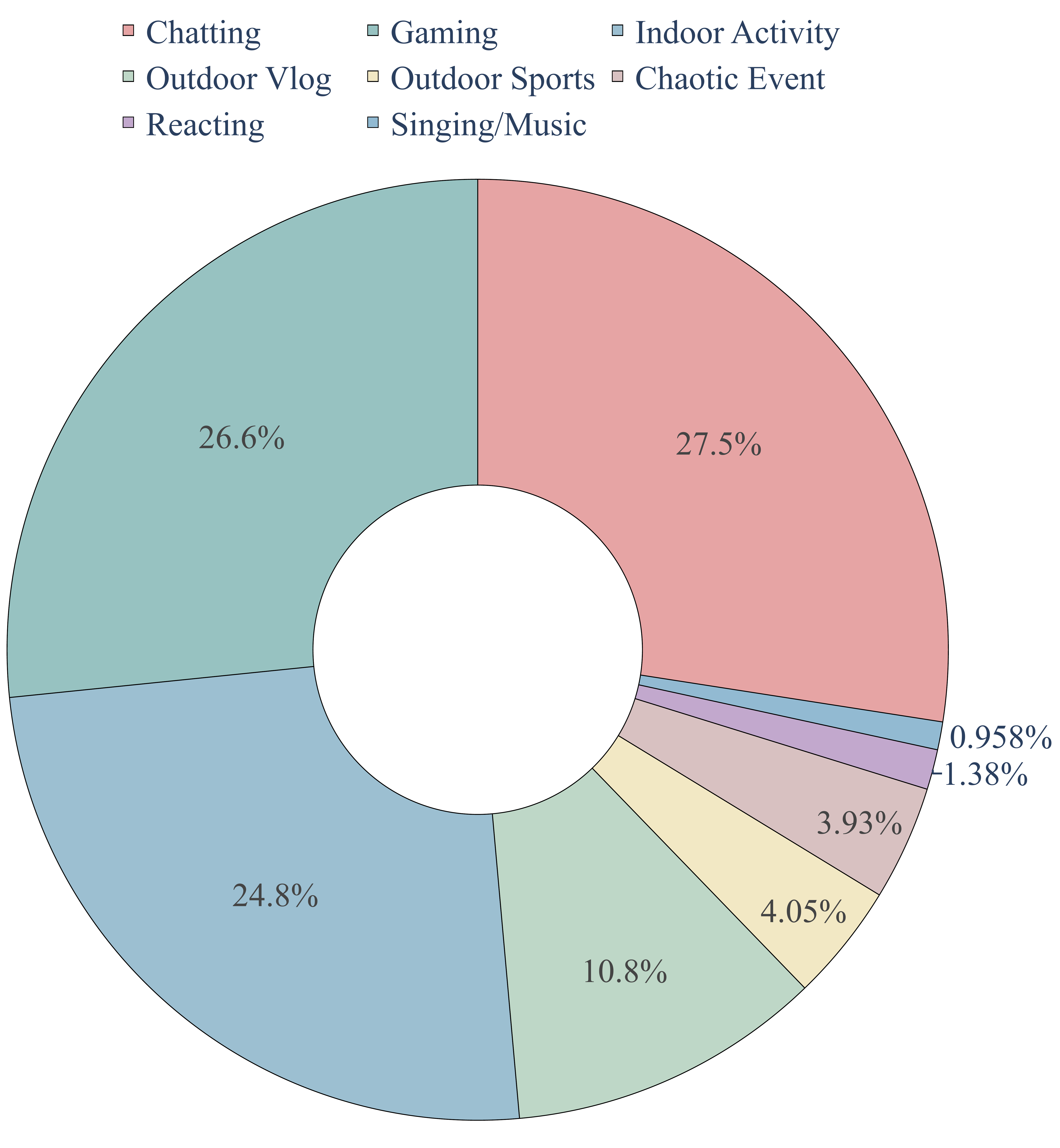}
\caption{Distribution of video clip domains.}
\label{fig:video-category}
\end{minipage}
\end{figure}

\begin{table}[t]
    \centering
    \small
    \caption{Statistics of the MM-Lifelong dataset.}
    \label{tab:stats}

    \setlength{\tabcolsep}{5pt}
    \renewcommand{\arraystretch}{1.1}

    \begin{tabular*}{\linewidth}{@{\extracolsep{\fill}}p{0.65\linewidth}r@{}}
        \toprule
        \textbf{Statistics} & \textbf{Number} \\
        \midrule
        Total Duration & 181.1 hours\\
        \midrule
        Total Questions & 1289 \\
        \quad Avg. Question Length & 26.79 words \\
        \quad Avg. Answer Length & 4.80 words \\
        \midrule
        Total Clue Intervals & 1810 (100\%) \\
        \quad Short ($<$90s) & 1039 (57.40\%)\\
        \quad Medium (90-540s) & 550 (30.39\%)\\
        \quad Long ($>$540s) & 221 (12.21\%)\\
        \quad Avg. Clue Duration & 362.26s \\
        \midrule
        Total Temporal Certificate & 1289 (100\%) \\
        \quad Short ($<$10m) & 500 (38.79\%)\\
        \quad Medium (10m-1h) & 395 (30.64\%)\\
        \quad Long (1h-10h) & 267 (20.71\%)\\
        \quad \textbf{Ultra-long ($>$10h)} & \textbf{127 (9.85\%)} \\
        \midrule
        Questions by Split & dev/val/test \\
        \quad Total & 266/623/400 \\
        \quad Gamer's Journey (Day) & 0/0/200 \\
        \quad Egocentric Life (Week) & 0/0/200 \\
        \quad Live Stream (Month) & 266/623/0 \\
        \bottomrule
    \end{tabular*}
\end{table}

As shown in Table~\ref{tab:stats}, MM-Lifelong comprises 181.1 hours of footage across three domains, with 1289 questions grounded by 1810 \emph{clue intervals}, the short evidence segments that directly verify each answer. Each question is also annotated with a \emph{temporal certificate}~\cite{mangalam2023egoschema}, the broader video region a human must inspect to reach the answer. Roughly $\mathbf{30\%}$ of questions have a temporal certificate exceeding one hour (267 in 1--10h, 127 beyond 10h), forcing multi-hour cross-session reasoning even though individual clue intervals remain short. The largest of the \textbf{11 question categories} are Counting (24.2\%), Causal Reasoning (14.6\%), and Entity Recognition (14.4\%). Among the \textbf{8 video clip domains}, Chatting (27.5\%), Gaming (26.6\%), and Indoor Activity (24.8\%) are the most common. Figures~\ref{fig:question-categroy} and~\ref{fig:video-category} show both distributions.

\begin{table*}[t]
\centering
\small
\caption{\textbf{Multimodal datasets across short-context, long-context, and lifelong settings.} \textbf{Max. Dur ($T_{dur}$)} is playback duration; \textbf{Max. Span ($T_{span}$)} is the underlying physical timeline. MM-Lifelong combines the \textit{Lifelong Horizon} ($T_{span} \gg T_{dur}$), with unobserved gaps over a nearly two-month span, and manual, clue-grounded annotation (\textbf{Clue}) of continuous audio-visual streams.}
\label{tab:comparison}
\setlength{\tabcolsep}{4pt}
\begin{tabular*}{\textwidth}{@{\extracolsep{\fill}}lccccccc@{}}
\toprule
\textbf{Dataset} & \textbf{Modalities} & \textbf{\#Samples} & \textbf{Max. Dur} & \textbf{Max. Span} & \textbf{Anno.} & \textbf{QA} & \textbf{Clue} \\
\midrule
\multicolumn{8}{l}{\textit{\textbf{I. Short-Context Multimodal Dataset}}} \\
MMMU~\cite{yue2024mmmu} & Image & 11.5k & 0 & 0 & M & 11.5k & \ding{55} \\
AIR-Bench~\cite{yang2024airbench} & Audio & 19k & 19.4s & 19.4s & A\&M & 19k & \ding{55} \\
OmniBench~\cite{li2024omnibench} & Audio+Image & 1.1k & 30s & 30s & A\&M & 1.1k & \ding{55} \\
MVBench~\cite{mvbench} & Video & 4.0k & 2.95m & 2.95m & A & 4.0k & \ding{55} \\
\midrule
\multicolumn{8}{l}{\textit{\textbf{II. Long-Context Multimodal Dataset}}} \\
EgoSchema~\cite{mangalam2023egoschema} & Video & 5.0k & 3.0m & 3.0m & A\&M & 5.0k & \ding{55} \\
Video-MME~\cite{videomme} & Video & 900 & 59.6m & 59.6m & M & 2.7k & \ding{55} \\
M3-Bench~\cite{long2025m3bench} & Video & 1020 & 57.5m & 57.5m & M & 4.9k & \ding{55} \\
CG-Bench~\cite{chen2024cg} & Video & 1219 & 1.75h & 1.75h & M & 12.2k & \checkmark \\
CG-AV-Counting~\cite{lu2025av-reasoner} & Audio+Video & 497 & 1.75h & 1.75h & M & 1.0k & \checkmark \\
\midrule
\multicolumn{8}{l}{\textit{\textbf{III. Lifelong Multimodal Dataset}}} \\
EgoLife~\cite{yang2025egolife} & Audio+Video & 6 & 51.9h & $\sim$7d & A\&M & 3.0k & \ding{55} \\
TeleEgo~\cite{yan2025teleego} & Audio+Video & 5 & 14.4h & $\sim$3d & A\&M & 3.3k & \ding{55} \\
\textbf{MM-Lifelong (Ours)} & Audio+Video & \textbf{3} & \textbf{105.6h} & \textbf{$\sim$51d} & \textbf{M} & \textbf{1.3k} & \checkmark \\
\bottomrule
\end{tabular*}
\end{table*}

\subsection{Benchmark Splits}
\label{subsec:splits}

We establish a split protocol that separates system development from evaluation under both temporal and domain shifts. Detailed statistics for each split are provided in Table~\ref{tab:stats}.

\textbf{Domain Generalization.} We reserve the Day-scale and Week-scale subsets exclusively for testing. Their unseen subjects and environments serve as held-out tests of out-of-distribution generalization.

\textbf{Temporal Partitioning.} For the Month-scale subset, we avoid a naive random split to prevent temporal data leakage. In lifelong streams, random sampling often places development and validation examples in close temporal proximity, allowing systems to exploit local environmental context. We sort all question-answer-clue triplets chronologically by their clue positions, assigning the first 30\% to the development split and the remaining 70\% to validation. The development split exposes labels for prompt, retrieval, and system design, while validation measures generalization to later, unseen segments of the same stream. The reported benchmark evaluation remains training-free and does not use the development examples for benchmark-specific model training.

\begin{table*}[t]
  \centering
  \footnotesize
  \setlength{\tabcolsep}{3pt}
  \renewcommand{\arraystretch}{1.06}
  \caption{\textbf{Performance comparison on val@month, test@week, and test@day of MM-Lifelong.} Score evaluates answers. Ref is temporal grounding at 300 seconds (Ref@300). Human reports accuracy in the Score columns. Models and agents are ordered by overall Score within each group. Bold and underline mark the best and second-best results within each group.}
  \label{tab:main_results}

  \begin{tabular*}{\textwidth}{@{\extracolsep{\fill}}lcrrrrrrrr@{}}
    \toprule
    & & \multicolumn{2}{c}{\textbf{Overall}} & \multicolumn{2}{c}{\textbf{Val@Month}} & \multicolumn{2}{c}{\textbf{Test@Week}} & \multicolumn{2}{c}{\textbf{Test@Day}} \\
    \cmidrule(lr){3-4} \cmidrule(lr){5-6} \cmidrule(lr){7-8} \cmidrule(lr){9-10}
    \textbf{System} & \textbf{Token cost} & \textbf{Score} & \textbf{Ref} & \textbf{Score} & \textbf{Ref} & \textbf{Score} & \textbf{Ref} & \textbf{Score} & \textbf{Ref} \\
    \midrule
    Human (Acc.) & -- & 88.45 & 38.43 & 82.70 & 33.50 & 96.10 & 42.40 & 98.70 & 49.80 \\
    \midrule
    \multicolumn{10}{l}{\textbf{End-to-End MLLMs}} \\
    \addlinespace[2pt]
    GPT-6.1-Sol~\citep{openai2026gpt61sol} & 1M & \textbf{31.65} & \textbf{27.29} & \textbf{28.48} & \textbf{22.20} & \textbf{36.80} & \textbf{31.16} & \textbf{36.40} & \textbf{39.28} \\
    Qwen3.5-9B~\citep{qwen2026qwen35} & 1M & \underline{21.72} & 2.58 & \underline{24.08} & 2.02 & 23.70 & 5.00 & 12.40 & 1.90 \\
    Qwen3.8-27B~\citep{qwen2026qwen38} & 1M & 21.45 & \underline{5.75} & 21.73 & \underline{4.80} & \underline{27.80} & \underline{9.45} & 14.20 & 5.01 \\
    Qwen3.5-35B-A3B~\citep{qwen2026qwen35} & 1M & 19.37 & 3.27 & 19.61 & 2.32 & 25.30 & 3.40 & 12.70 & \underline{6.13} \\
    Gemma-4-31B-IT~\citep{gemmateam2026gemma4} & 256K & 19.34 & 2.74 & 20.35 & 1.99 & 22.00 & 7.03 & 13.50 & 0.81 \\
    VideoChat3-4B~\citep{li2026videochat3} & 104K & 18.26 & 0.34 & 18.68 & 0.26 & 24.00 & 0.73 & 11.20 & 0.17 \\
    Nemotron-3-Nano-Omni-30B-A3B~\citep{nvidia2026nemotron3nanoomni} & 256K & 18.08 & 0.16 & 18.17 & 0.01 & 23.60 & 0.52 & 12.30 & 0.25 \\
    GLM-4.6V-106B~\citep{vteam2025glm} & 256K & 17.67 & 3.72 & 17.66 & 3.68 & 22.90 & 6.51 & 12.50 & 1.03 \\
    Gemma-4-12B-IT~\citep{gemmateam2026gemma4} & 256K & 17.09 & 0.43 & 18.56 & 0.19 & 20.70 & 1.21 & 8.90 & 0.39 \\
    InternVideo3-8B-Instruct~\citep{yan2026internvideo3} & 128K & 16.91 & 0.20 & 17.85 & 0.08 & 19.40 & 0.50 & 11.50 & 0.25 \\
    Ministral-3-14B~\citep{liu2026ministral3} & 256K & 16.85 & 1.80 & 17.18 & 0.82 & 18.20 & 5.67 & 14.50 & 0.97 \\
    Eagle-2.5-8B~\citep{chen2025eagle} & 1M & 16.58 & 0.55 & 16.53 & 0.03 & 19.50 & 1.69 & 13.80 & 1.01 \\
    GLM-4.6V-Flash-9B~\citep{vteam2025glm} & 256K & 15.95 & 2.55 & 14.25 & 2.24 & 22.00 & 5.21 & \underline{15.20} & 0.88 \\
    Video-XL-2~\citep{qin2025video} & 128K & 11.59 & 0.70 & 13.00 & 0.75 & 12.20 & 0.51 & 6.60 & 0.72 \\
    Qwen3-Omni-30B-A3B~\citep{xu2025qwen3omni} & 32K & 11.40 & 0.26 & 12.20 & 0.25 & 8.20 & 0.54 & 12.10 & 0.02 \\
    VideoXL-Pro-3B~\citep{liu2025videoxlpro} & 32K & 8.54 & 0.12 & 10.79 & 0.00 & 5.90 & 0.63 & 4.20 & 0.00 \\
    \midrule
    \multicolumn{10}{l}{\textbf{Agents}} \\
    \addlinespace[2pt]
    Codex (GPT-6.1-Sol)~\citep{openai2026codex} & -- & \textbf{42.33} & \textbf{52.32} & \textbf{38.33} & \textbf{51.16} & \textbf{32.80} & \textbf{43.77} & \textbf{64.30} & \textbf{64.46} \\
    VideoITG~\citep{wang2025videoitg} & -- & \underline{19.75} & 1.95 & \underline{20.22} & 1.17 & \underline{21.50} & 4.05 & \underline{16.50} & 2.28 \\
    WorldMM~\citep{yeo2025worldmm} & -- & 13.80 & 5.52 & 13.58 & \underline{5.54} & 16.20 & 5.05 & 12.10 & 5.92 \\
    VideoZoomer~\citep{ding2025videozoomer} & -- & 13.27 & 0.37 & 13.26 & 0.04 & 16.00 & 1.16 & 10.60 & 0.63 \\
    ParaVT~\citep{yang2026paravt} & -- & 12.77 & 0.16 & 14.09 & 0.00 & 12.30 & 0.82 & 9.10 & 0.00 \\
    SAGE~\citep{jain2025sage} & -- & 11.09 & 4.03 & 11.20 & 3.80 & 9.30 & 3.22 & 12.50 & 5.53 \\
    LongVideoAgent~\citep{liu2025longvideoagent} & -- & 10.09 & 3.46 & 9.60 & 3.31 & 14.90 & 4.50 & 6.80 & 2.86 \\
    TaskMem~\citep{zou2026taskmem} & -- & 7.92 & \underline{6.51} & 7.93 & 5.39 & 6.10 & \underline{5.68} & 9.70 & \underline{10.85} \\
    \bottomrule
  \end{tabular*}
\end{table*}

\subsection{Evaluation Protocol}
\label{sec:eval}

We benchmark models on answer quality and clue-level temporal localization.

\textbf{Building the answer rubrics.} A single reference answer cannot cover all valid phrasings or distinguish a partially correct response from a wrong one. We therefore build a separate rubric for every question. Human annotators first specify the facts that a correct answer must contain, valid alternatives, acceptable numerical or temporal ranges, and conditions that make an answer incorrect. GPT-5.6-sol then examines this specification, the annotated video evidence, and example responses from diverse MLLMs and agents. It adds only evidence-supported paraphrases and aliases and defines what scores from 0 to 5 mean for that question. The example responses help reveal common phrasings and errors, but do not determine the correct answer. We freeze all rubrics before scoring the benchmark.

\textbf{Scoring model answers.} For each model, we evaluate each prediction against the corresponding frozen rubric. GPT-5.6-luna~\citep{openai2026luna} serves as the rubric-based verifier, assessing the requested facts, omissions, and contradictions using the question, reference answer, and scoring criteria. Each prediction receives $s\in\{0,1,2,3,4,5\}$, with missing predictions assigned zero. For a split with $N$ questions, we report $100\sum_i s_i/(5N)$. This preserves partial correctness while applying the same question-specific criteria to every model. Appendix~\ref{app:exp:eval} provides the grading template and a worked example.

\textbf{Reference Grounding (Ref@N).} Exact interval overlap is overly sensitive when short clues must be localized within streams lasting many hours. We divide the video timeline into non-overlapping bins of $N$ seconds and mark the bins touched by the predicted and reference intervals. Ref@N is the intersection-over-union of these two bin sets, $\text{Ref@N}(P,G)=|P\cap G|/|P\cup G|\times100$. We use $N{=}300\text{s}$ by default. Smaller bins require more precise localization, whereas larger bins allow greater temporal tolerance.

\subsection{Comparison with Existing Multimodal Datasets}
\label{subsec:comparison}
Table~\ref{tab:comparison} positions MM-Lifelong against representative multimodal datasets across short-context, long-context, and lifelong settings. Within the lifelong category, MM-Lifelong differs from the closest prior efforts (EgoLife~\cite{yang2025egolife}, TeleEgo~\cite{yan2025teleego}) along four dimensions: \textit{temporal scope} (the Month subset reaches month-scale $T_{span}$), \textit{annotation} (manual clue-grounded throughout), \textit{evaluation metric} (Ref@N for sparse-clue grounding), and \textit{evaluation protocol} (a chronological dev/val split at the Month scale, with Day and Week as held-out cross-domain tests). Combined with multi-domain coverage across gamer, wearer, and streamer subjects, these properties make MM-Lifelong the first benchmark to operationalize $T_{span} \gg T_{dur}$ at the month scale with manual, clue-grounded annotations.

\section{Evaluation Results}
\label{sec:benchmark}

\subsection{Evaluation Setup}
\label{subsec:baselines}
We evaluate MM-Lifelong in a \textit{training-free} setting, comparing existing systems within two technical routes without benchmark-specific training.

\textbf{End-to-end MLLMs} answer directly from sampled video frames. We include general-purpose models such as \textit{GPT-6.1-Sol} and \textit{Qwen3.8-27B}~\citep{qwen2026qwen38}, and video-specialized models such as \textit{Eagle-2.5}~\citep{chen2025eagle}. Grounding intervals are predicted directly. 

\textbf{Agentic methods} access the stream through selection, retrieval, memory, or iterative inspection. Examples include \textit{Codex}~\citep{openai2026codex}, \textit{VideoITG}~\citep{wang2025videoitg}, and \textit{WorldMM}~\citep{yeo2025worldmm}. Grounding uses reported intervals, selected frames, or tool traces.

Table~\ref{tab:main_results} pairs answer scores with Ref@300. Appendices~\ref{app:exp} and~\ref{app:compute} give the protocols and compute details.

\subsection{Results and Analysis}
\label{subsec:results}

\textbf{Answer scores vary with input budget.}
GPT-6.1-Sol scores 31.57 at 32K tokens, peaks at 32.98 at 256K, and scores 31.65 at 1M (Figure~\ref{fig:overall}(b)). Over the same 32K-to-1M range, Qwen3.5-9B gains 4.36 points and Eagle-2.5-8B gains 0.49 points. Eagle-2.5-8B reaches its best score at 128K. The budget that yields the highest answer score therefore varies across models.

\textbf{Answer and grounding scores reveal different strengths.}
Codex (GPT-6.1-Sol) leads the agents on both metrics, with an overall answer score of 42.33 and a Ref@300 of 52.32 (Table~\ref{tab:main_results}). Its answer scores vary by split, from 32.80 on Week and 38.33 on Month to 64.30 on Day. In the end-to-end setting at 1M, GPT-6.1-Sol scores 31.65 for answers and 27.29 for grounding. Other agents rank differently across the two metrics. VideoITG has the second-highest overall answer score among agents (19.75), with a Ref@300 of 1.95. TaskMem places second on grounding (6.51) and has the lowest agent answer score (7.92).

\subsection{Error Analysis}
\label{subsec:error-analysis}

\begin{figure}[htb]
  \centering
  \includegraphics[width=\linewidth]{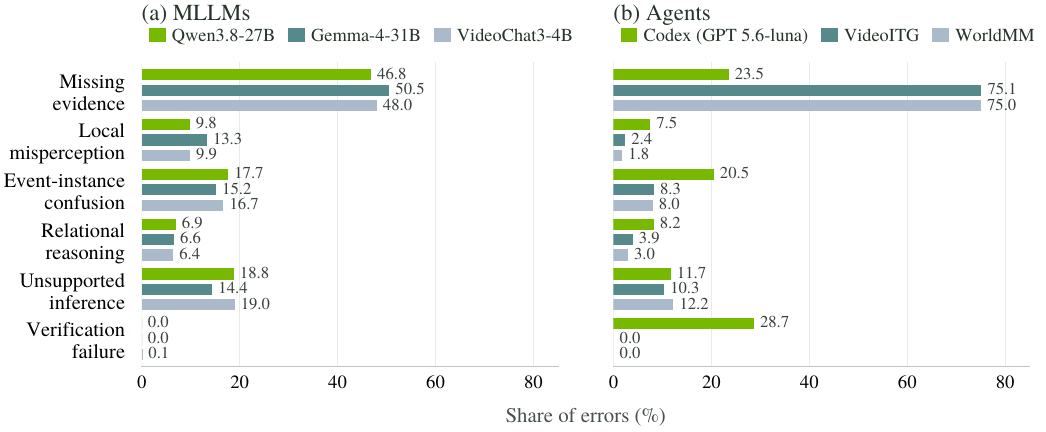}
  \caption{Error breakdown by system for predictions scoring below 4. Each prediction has one primary error label.}
  \label{fig:error-analysis}
\end{figure}

We analyze the errors of representative MLLMs and agents to identify what limits lifelong understanding. Figure~\ref{fig:error-analysis} highlights recurring failures in evidence retention, event identification, and verification.

\paragraph{Loss of answer-relevant evidence.}
A scene-level summary can remain accurate while losing the details needed to reconstruct what happened. Missing evidence is the largest category, accounting for 46.8--50.5\% of MLLM errors, 75.1\% of VideoITG errors, and 75.0\% of WorldMM errors. Inspected sampling records show brief state changes disappearing while surrounding scenes remain available. This challenges scene similarity as a compression criterion: visually redundant moments can contain distinct facts. Such gaps can be hard to detect from a summary that remains internally coherent. Future memory systems should retain state changes and source links for query-time recovery. They should also track observation gaps, so that a missing record is not mistaken for evidence that an event never occurred.

\paragraph{Confusion between event instances.}
Semantic relevance does not establish that a retrieved event is the one being queried. Among MLLM errors, instance confusion accounts for 15.2--17.7\%, versus 6.4--6.9\% for relational reasoning. Inspected cases show locally accurate observations supplying the wrong premise when bound to another occurrence or role. Describing that occurrence in greater detail can strengthen a misleading match without establishing its identity. Further reasoning may then elaborate the error rather than correct it. This motivates separating candidate retrieval from instance verification: preserve event identities, temporal anchors, and provenance, and verify these bindings before combining events. These bindings need to survive summarization as well as retrieval.

\paragraph{Failure to resolve conflicting evidence.}
Verification can fail even after uncertainty has been explicitly recognized. Verification failure accounts for 28.7\% of Codex's errors, and unsupported inference for 11.7\%. Inspected traces show conflicting reports from identical frames followed by a definite answer. The final choice leaves the contradiction unresolved. Because these reports share the same observations, their repetition cannot be counted as independent support. Re-querying the same frames may refine a reading, but cannot supply an observation missing from those frames. Controllers should retain competing explanations, inspect observations that distinguish them, and base stopping on evidential support rather than the completion of another tool call. Unresolved conflicts must remain attached to the claim through final answer generation.

\vspace{-3mm}
\subsection{Verifier Agreement}
\label{subsec:verifier-human}

Three volunteers used the frozen rubrics to grade 400 responses, 100 from each of two MLLMs and two agents. Gemini 4 Argon and GPT-6.1-Sol re-graded all responses from the archived evaluation with the same frozen rubrics, blind to system identities and original scores. Table~\ref{tab:verifier-human} summarizes both studies.

\begin{table}[!htb]
  \centering
  \caption{\textbf{Agreement between graders.} $\kappa$ is quadratic-weighted. MAE uses the 0--5 score scale. Verifier--human uses adjudicated consensus. Human--human averages three rater pairs.}
  \label{tab:verifier-human}
  \small
  \setlength{\tabcolsep}{3pt}
  \begin{tabular*}{\linewidth}{@{\extracolsep{\fill}}lccc@{}}
    \toprule
    \textbf{Comparison} & $\boldsymbol{\kappa}$ & \textbf{MAE} & \textbf{Pearson} $\boldsymbol{r}$ \\
    \midrule
    \multicolumn{4}{@{}l}{\textit{Human grading}} \\
    Verifier vs.\ human consensus & 0.63 & 0.69 & 0.75 \\
    Human vs.\ human & 0.67 & 0.58 & 0.81 \\
    \midrule
    \multicolumn{4}{@{}l}{\textit{Model grading}} \\
    Verifier vs.\ Gemini 4 Argon & 0.89 & 0.22 & 0.90 \\
    Verifier vs.\ GPT-6.1-Sol & 0.90 & 0.22 & 0.90 \\
    Gemini 4 Argon vs.\ GPT-6.1-Sol & 0.95 & 0.11 & 0.95 \\
    \bottomrule
  \end{tabular*}
\end{table}
\FloatBarrier

\vspace{-3mm}
\section{Conclusion}
\label{sec:conclusion}
\vspace{-2mm}

We introduce \textbf{MM-Lifelong}, a benchmark for multimodal understanding over subject-specific histories at Day, Week, and Month scales. It distinguishes observational duration from physical temporal span and connects questions to manually annotated evidence. Our evaluation reveals uneven gains from larger input budgets and differing system strengths in answering and temporal grounding. Error analysis highlights the need to retain event-specific details, maintain correct event identities, and resolve conflicting observations. These findings motivate memory systems that preserve explicit links between answer claims and supporting evidence. The current benchmark covers one primary subject per scale and a maximum span of 51 days. Future work should broaden subject coverage, extend temporal horizons, and explicitly evaluate reasoning about unobserved intervals. We release the benchmark, scoring rubrics, and baselines to support these efforts.

\bibliographystyle{icml2026}
\bibliography{references}

@inproceedings{chen2024cg,
  title={{CG-Bench}: Clue-grounded Question Answering Benchmark for Long Video Understanding},
  author={Chen, Guo and Liu, Yicheng and Huang, Yifei and Pei, Baoqi and Xu, Jilan and He, Yuping and Lu, Tong and Wang, Yali and Wang, Limin},
  booktitle={International Conference on Learning Representations},
  year={2025},
  url={https://proceedings.iclr.cc/paper_files/paper/2025/hash/70fa5df8e3300dc30bf19bee44a56155-Abstract-Conference.html}
}

@article{huang2025vinci,
  title={Vinci: A real-time smart assistant based on egocentric vision-language model for portable devices},
  author={Huang, Yifei and Xu, Jilan and Pei, Baoqi and Yang, Lijin and Zhang, Mingfang and He, Yuping and Chen, Guo and Chen, Xinyuan and Wang, Yaohui and Nie, Zheng and others},
  journal={Proceedings of the ACM on Interactive, Mobile, Wearable and Ubiquitous Technologies},
  volume={9},
  number={3},
  pages={1--33},
  year={2025},
  doi={10.1145/3749513},
  publisher={ACM New York, NY, USA}
}

@inproceedings{pei2025egothinker,
  title={{EgoThinker}: Unveiling Egocentric Reasoning with Spatio-Temporal {CoT}},
  author={Pei, Baoqi and Huang, Yifei and Xu, Jilan and He, Yuping and Chen, Guo and Wu, Fei and Pang, Jiangmiao and Qiao, Yu},
  booktitle={Advances in Neural Information Processing Systems},
  volume={38},
  year={2025},
  url={https://proceedings.neurips.cc/paper_files/paper/2025/hash/3e9412a9c1d93810ef3ef7825115016b-Abstract-Conference.html}
}

@inproceedings{mvbench,
	title        = {Mvbench: A comprehensive multi-modal video understanding benchmark},
	author       = {Li, Kunchang and Wang, Yali and He, Yinan and Li, Yizhuo and Wang, Yi and Liu, Yi and Wang, Zun and Xu, Jilan and Chen, Guo and Luo, Ping and others},
	year         = 2024,
	booktitle    = {CVPR},
	pages        = {22195--22206}
}

@inproceedings{li2024omnibench,
  title={{OmniBench}: Towards The Future of Universal Omni-Language Models},
  author={Li, Yizhi and Zhang, Ge and Ma, Yinghao and Yuan, Ruibin and others},
  booktitle={Advances in Neural Information Processing Systems},
  volume={38},
  year={2025},
  url={https://proceedings.neurips.cc/paper_files/paper/2025/hash/2c7c4a12a9dcf7dace9896d08154c705-Abstract-Datasets_and_Benchmarks_Track.html}
}

@inproceedings{huang2018predicting,
	title        = {Predicting gaze in egocentric video by learning task-dependent attention transition},
	author       = {Huang, Yifei and Cai, Minjie and Li, Zhenqiang and Sato, Yoichi},
	year         = 2018,
	booktitle    = {Proceedings of the ECCV (ECCV)},
	pages        = {754--769}
}

@inproceedings{ego4d,
	title        = {Ego4d: Around the world in 3,000 hours of egocentric video},
	author       = {Grauman, Kristen and Westbury, Andrew and Byrne, Eugene and Chavis, Zachary and Furnari, Antonino and Girdhar, Rohit and Hamburger, Jackson and Jiang, Hao and Liu, Miao and Liu, Xingyu and others},
	year         = 2022,
	booktitle    = {CVPR},
	pages        = {18995--19012}
}

@inproceedings{huang2020improving,
	title        = {Improving action segmentation via graph-based temporal reasoning},
	author       = {Huang, Yifei and Sugano, Yusuke and Sato, Yoichi},
	year         = 2020,
	booktitle    = {CVPR},
	pages        = {14024--14034}
}

@inproceedings{mae,
	title        = {Masked Autoencoders Are Scalable Vision Learners},
	author       = {Kaiming He and Xinlei Chen and Saining Xie and Yanghao Li and Piotr Doll{\'{a}}r and Ross B. Girshick},
	year         = 2022,
	booktitle    = {CVPR},
	pages        = {15979--15988}
}

@inproceedings{videomme,
  title={{Video-MME}: The First-Ever Comprehensive Evaluation Benchmark of Multi-modal {LLMs} in Video Analysis},
  author={Fu, Chaoyou and Dai, Yuhan and Luo, Yongdong and Li, Lei and Ren, Shuhuai and Zhang, Renrui and Wang, Zihan and Zhou, Chenyu and Shen, Yunhang and Zhang, Mengdan and Chen, Peixian and Li, Yanwei and Lin, Shaohui and Zhao, Sirui and Li, Ke and Xu, Tong and Zheng, Xiawu and Chen, Enhong and Shan, Caifeng and He, Ran and Sun, Xing},
  booktitle={Proceedings of the IEEE/CVF Conference on Computer Vision and Pattern Recognition},
  pages={24108--24118},
  year={2025},
  url={https://openaccess.thecvf.com/content/CVPR2025/html/Fu_Video-MME_The_First-Ever_Comprehensive_Evaluation_Benchmark_of_Multi-modal_LLMs_in_CVPR_2025_paper.html}
}

@inproceedings{egoexolearn,
	title        = {EgoExoLearn: A Dataset for Bridging Asynchronous Ego- and Exo-centric View of Procedural Activities in Real World},
	author       = {Huang, Yifei and Chen, Guo and Xu, Jilan and Zhang, Mingfang and Yang, Lijin and Pei, Baoqi and Zhang, Hongjie and Lu, Dong and Wang, Yali and Wang, Limin and Qiao, Yu},
	year         = 2024,
	booktitle    = {CVPR}
}

@article{gemini,
	title        = {Gemini: {A} Family of Highly Capable Multimodal Models},
	author       = {Rohan Anil and Sebastian Borgeaud and Yonghui Wu and Jean{-}Baptiste Alayrac and Jiahui Yu and Radu Soricut and Johan Schalkwyk and Andrew M. Dai and Anja Hauth and Katie Millican and David Silver and Slav Petrov and Melvin Johnson and Ioannis Antonoglou and Julian Schrittwieser and Amelia Glaese and Jilin Chen and Emily Pitler and Timothy P. Lillicrap and Angeliki Lazaridou and Orhan Firat and James Molloy and Michael Isard and Paul Ronald Barham and Tom Hennigan and Benjamin Lee and Fabio Viola and Malcolm Reynolds and Yuanzhong Xu and Ryan Doherty and Eli Collins and Clemens Meyer and Eliza Rutherford and Erica Moreira and Kareem Ayoub and Megha Goel and George Tucker and Enrique Piqueras and Maxim Krikun and Iain Barr and Nikolay Savinov and Ivo Danihelka and Becca Roelofs and Ana{\"{\i}}s White and Anders Andreassen and Tamara von Glehn and Lakshman Yagati and Mehran Kazemi and Lucas Gonzalez and Misha Khalman and Jakub Sygnowski and et al.},
	year         = 2023,
	journal      = {CoRR},
	volume       = {abs/2312.11805}
}

@inproceedings{wu2024longvideobench,
  title={{LongVideoBench}: A Benchmark for Long-context Interleaved Video-Language Understanding},
  author={Wu, Haoning and Li, Dongxu and Chen, Bei and Li, Junnan},
  booktitle={Advances in Neural Information Processing Systems},
  volume={37},
  year={2024},
  url={https://proceedings.neurips.cc/paper_files/paper/2024/hash/329ad516cf7a6ac306f29882e9c77558-Abstract-Datasets_and_Benchmarks_Track.html}
}

@article{ning2023videobench,
  title={{Video-Bench}: A Comprehensive Benchmark and Toolkit for Evaluating Video-Based Large Language Models},
  author={Ning, Munan and Zhu, Bin and Xie, Yujia and Lin, Bin and Cui, Jiaxi and Yuan, Lu and Chen, Dongdong and Yuan, Li},
  journal={Computational Visual Media},
  volume={12},
  number={1},
  pages={71--84},
  year={2026},
  doi={10.26599/CVM.2025.9450516},
  url={https://www.sciopen.com/article/10.26599/CVM.2025.9450516}
}

@inproceedings{mangalam2023egoschema,
  author       = {Karttikeya Mangalam and
                  Raiymbek Akshulakov and
                  Jitendra Malik},
  title        = {EgoSchema: {A} Diagnostic Benchmark for Very Long-form Video Language
                  Understanding},
  booktitle    = {NeurIPS},
  year         = {2023}
}

@inproceedings{chen2025eagle,
  title={{Eagle 2.5}: Boosting Long-Context Post-Training for Frontier Vision-Language Models},
  author={Chen, Guo and Li, Zhiqi and Wang, Shihao and Jiang, Jindong and Liu, Yicheng and Lu, Lidong and Huang, De-An and Byeon, Wonmin and Le, Matthieu and Ehrlich, Max and others},
  booktitle={Advances in Neural Information Processing Systems},
  volume={38},
  year={2025},
  url={https://proceedings.neurips.cc/paper_files/paper/2025/hash/833295cb9278a3ba973842a94ea68e3c-Abstract-Conference.html}
}

@misc{google_2025_deepsearch,
	title        = {Gemini Deep Research Demo | Using AI to learn new topics in depth},
	author       = {{Google}},
	year         = 2025,
	howpublished = {YouTube video},
	url          = {https://www.youtube.com/watch?v=buwMJxvW7wI}
}

@inproceedings{chen2025lvagent,
	title        = {LVAgent: Long Video Understanding by Multi-Round Dynamical Collaboration of MLLM Agents},
	author       = {Chen, Boyu and Yue, Zhengrong and Chen, Siran and Wang, Zikang and Liu, Yang and Li, Peng and Wang, Yali},
	year         = 2025,
	booktitle    = {Proceedings of the IEEE/CVF International Conference on Computer Vision},
	pages        = {20237--20246},
	url          = {https://openaccess.thecvf.com/content/ICCV2025/html/Chen_LVAgent_Long_Video_Understanding_by_Multi-Round_Dynamical_Collaboration_of_MLLM_ICCV_2025_paper.html}
}

@inproceedings{yang2025egolife,
  title={{EgoLife}: Towards Egocentric Life Assistant},
  author={Yang, Jingkang and Liu, Shuai and Guo, Hongming and Dong, Yuhao and Zhang, Xiamengwei and Zhang, Sicheng and Wang, Pengyun and Zhou, Zitang and Xie, Binzhu and Wang, Ziyue and others},
  booktitle={Proceedings of the IEEE/CVF Conference on Computer Vision and Pattern Recognition},
  pages={28885--28900},
  year={2025},
  url={https://openaccess.thecvf.com/content/CVPR2025/html/Yang_EgoLife_Towards_Egocentric_Life_Assistant_CVPR_2025_paper.html}
}

@article{yang2025qwen3,
  title={Qwen3 Technical Report},
  author={Yang, An and Li, Anfeng and Yang, Baosong and Zhang, Beichen and Hui, Binyuan and Zheng, Bo and Yu, Bowen and Gao, Chang and Huang, Chengen and Lv, Chenxu and others},
  journal={arXiv preprint arXiv:2505.09388},
  year={2025}
}

@inproceedings{yue2024mmmu,
  title={Mmmu: A massive multi-discipline multimodal understanding and reasoning benchmark for expert agi},
  author={Yue, Xiang and Ni, Yuansheng and Zhang, Kai and Zheng, Tianyu and Liu, Ruoqi and Zhang, Ge and Stevens, Samuel and Jiang, Dongfu and Ren, Weiming and Sun, Yuxuan and others},
  booktitle={Proceedings of the IEEE/CVF Conference on Computer Vision and Pattern Recognition},
  pages={9556--9567},
  year={2024}
}

@inproceedings{liu2024mmbench,
  title={Mmbench: Is your multi-modal model an all-around player?},
  author={Liu, Yuan and Duan, Haodong and Zhang, Yuanhan and Li, Bo and Zhang, Songyang and Zhao, Wangbo and Yuan, Yike and Wang, Jiaqi and He, Conghui and Liu, Ziwei and others},
  booktitle={European conference on computer vision},
  pages={216--233},
  year={2024},
  organization={Springer}
}

@inproceedings{peng2025mvu-eval,
  title={MVU-Eval: Towards Multi-Video Understanding Evaluation for Multimodal LLMs},
  author={Peng, Tianhao and Wang, Haochen and Zhang, Yuanxing and Wang, Noah and others},
  booktitle={Advances in Neural Information Processing Systems},
  volume={38},
  year={2025},
  url={https://proceedings.neurips.cc/paper_files/paper/2025/hash/4dbb159f2c68b359b3b6ed7dd039481c-Abstract-Datasets_and_Benchmarks_Track.html}
}

@article{zhu2025cvbench,
  title={Cvbench: Evaluating cross-video synergies for complex multimodal understanding and reasoning},
  author={Zhu, Nannan and Dong, Yonghao and Wang, Teng and Li, Xueqian and Deng, Shengjun and Wang, Yijia and Hong, Zheng and Geng, Tiantian and Niu, Guo and Huang, Hanyan and others},
  journal={arXiv preprint arXiv:2508.19542},
  year={2025}
}

@article{kuratov2024babilong,
  title={Babilong: Testing the limits of llms with long context reasoning-in-a-haystack},
  author={Kuratov, Yury and Bulatov, Aydar and Anokhin, Petr and Rodkin, Ivan and Sorokin, Dmitry and Sorokin, Artyom and Burtsev, Mikhail},
  journal={Advances in Neural Information Processing Systems},
  volume={37},
  pages={106519--106554},
  year={2024}
}

@inproceedings{bai2024longbench,
  title={Longbench: A bilingual, multitask benchmark for long context understanding},
  author={Bai, Yushi and Lv, Xin and Zhang, Jiajie and Lyu, Hongchang and Tang, Jiankai and Huang, Zhidian and Du, Zhengxiao and Liu, Xiao and Zeng, Aohan and Hou, Lei and others},
  booktitle={Proceedings of the 62nd annual meeting of the association for computational linguistics (volume 1: Long papers)},
  pages={3119--3137},
  year={2024}
}

@inproceedings{bei2026mem-gallery,
  title={Mem-Gallery: Benchmarking Multimodal Long-Term Conversational Memory for MLLM Agents},
  author={Bei, Yuanchen and Wei, Tianxin and Ning, Xuying and Zhao, Yanjun and Liu, Zhining and Lin, Xiao and Zhu, Yada and Hamann, Hendrik and He, Jingrui and Tong, Hanghang},
  booktitle={Proceedings of the 64th Annual Meeting of the Association for Computational Linguistics},
  pages={40750--40784},
  year={2026},
  doi={10.18653/v1/2026.acl-long.1892},
  url={https://aclanthology.org/2026.acl-long.1892/}
}

@inproceedings{yang2025svbench,
  title={{SVBench}: A Benchmark with Temporal Multi-Turn Dialogues for Streaming Video Understanding},
  author={Yang, Zhenyu and Hu, Yuhang and Du, Zemin and Xue, Dizhan and Qian, Shengsheng and Wu, Jiahong and Yang, Fan and Dong, Weiming and Xu, Changsheng},
  booktitle={International Conference on Learning Representations},
  year={2025},
  url={https://proceedings.iclr.cc/paper_files/paper/2025/hash/693ce820ea7e0f0c70a2a833cbdf7ccc-Abstract-Conference.html}
}

@article{liu2024deepseekv3,
  title={{DeepSeek-V3} Technical Report},
  author={{DeepSeek-AI} and Liu, Aixin and Feng, Bei and Xue, Bing and Wang, Bingxuan and Wu, Bochao and Lu, Chengda and Zhao, Chenggang and Deng, Chengqi and Zhang, Chenyu and others},
  journal={arXiv preprint arXiv:2412.19437},
  year={2024},
  url={https://arxiv.org/abs/2412.19437}
}

@inproceedings{cheng2026deepseek-engram,
  title={Conditional Memory via Scalable Lookup: A New Axis of Sparsity for Large Language Models},
  author={Cheng, Xin and Zeng, Wangding and Dai, Damai and Chen, Qinyu and Wang, Bingxuan and Xie, Zhenda and Huang, Kezhao and Yu, Xingkai and Hao, Zhewen and Zhang, Han and Li, Yu-Kun and Zhang, Huishuai and Zhao, Dongyan and Liang, Wenfeng},
  booktitle={Proceedings of the 64th Annual Meeting of the Association for Computational Linguistics},
  pages={4968--4990},
  year={2026},
  doi={10.18653/v1/2026.acl-long.226},
  url={https://aclanthology.org/2026.acl-long.226/}
}

@article{blakeman2025nemotron-h,
  title={Nemotron-h: A family of accurate and efficient hybrid mamba-transformer models},
  author={Blakeman, Aaron and Basant, Aarti and Khattar, Abhinav and Renduchintala, Adithya and Bercovich, Akhiad and Ficek, Aleksander and Bjorlin, Alexis and Taghibakhshi, Ali and Deshmukh, Amala Sanjay and Mahabaleshwarkar, Ameya Sunil and others},
  journal={arXiv preprint arXiv:2504.03624},
  year={2025}
}

@inproceedings{xiao2023streamingllm,
  title={Efficient Streaming Language Models with Attention Sinks},
  author={Xiao, Guangxuan and Tian, Yuandong and Chen, Beidi and Han, Song and Lewis, Mike},
  booktitle={International Conference on Learning Representations},
  year={2024},
  url={https://proceedings.iclr.cc/paper_files/paper/2024/hash/5e5fd18f863cbe6d8ae392a93fd271c9-Abstract-Conference.html}
}

@inproceedings{zhang2025dvd,
  title={Deep Video Discovery: Agentic Search with Tool Use for Long-form Video Understanding},
  author={Zhang, Xiaoyi and Jia, Zhaoyang and Guo, Zongyu and Li, Jiahao and Li, Bin and Li, Houqiang and Lu, Yan},
  booktitle={Advances in Neural Information Processing Systems},
  volume={38},
  year={2025},
  url={https://proceedings.neurips.cc/paper_files/paper/2025/hash/8190b210e9808e54ee16263b673a847d-Abstract-Conference.html}
}

@inproceedings{faysse2024colpali,
  title={{ColPali}: Efficient Document Retrieval with Vision Language Models},
  author={Faysse, Manuel and Sibille, Hugues and Wu, Tony and Omrani, Bilel and Viaud, Gautier and Hudelot, C{\'e}line and Colombo, Pierre},
  booktitle={International Conference on Learning Representations},
  year={2025},
  url={https://proceedings.iclr.cc/paper_files/paper/2025/hash/99e9e141aafc314f76b0ca3dd66898b3-Abstract-Conference.html}
}

@article{wan2025mmgraphrag,
  title={Mmgraphrag: Bridging vision and language with interpretable multimodal knowledge graphs},
  author={Wan, Xueyao and Yu, Hang},
  journal={arXiv preprint arXiv:2507.20804},
  year={2025}
}

@inproceedings{chhikara2025mem0,
  title={{Mem0}: Building Production-Ready {AI} Agents with Scalable Long-Term Memory},
  author={Chhikara, Prateek and Khant, Dev and Aryan, Saket and Singh, Taranjeet and Yadav, Deshraj},
  booktitle={ECAI 2025},
  pages={2993--3000},
  year={2025},
  doi={10.3233/FAIA251160}
}

@inproceedings{he2025egoexobench,
  title={{EgoExoBench}: A Benchmark for First- and Third-person View Video Understanding in {MLLMs}},
  author={He, Yuping and Huang, Yifei and Chen, Guo and Pei, Baoqi and Xu, Jilan and Lu, Tong and Pang, Jiangmiao},
  booktitle={Advances in Neural Information Processing Systems},
  volume={38},
  year={2025},
  url={https://proceedings.neurips.cc/paper_files/paper/2025/hash/a974bceb4d1af7d46673d1876482bd4f-Abstract-Datasets_and_Benchmarks_Track.html}
}

@inproceedings{lu2025av-reasoner,
  title={{AV-Reasoner}: Improving and Benchmarking Clue-Grounded Audio-Visual Counting for {MLLMs}},
  author={Lu, Lidong and Chen, Guo and Wei, Zhu and Li, Zhiqi and Liu, Yicheng and Lu, Tong},
  booktitle={Proceedings of the IEEE/CVF Conference on Computer Vision and Pattern Recognition},
  pages={33477--33487},
  year={2026},
  url={https://openaccess.thecvf.com/CVPR2026?day=2026-06-07}
}

@inproceedings{yang2024airbench,
  title={{AIR-Bench}: Benchmarking Large Audio-Language Models via Generative Comprehension},
  author={Yang, Qian and Xu, Jin and Liu, Wenrui and Chu, Yunfei and Jiang, Ziyue and Zhou, Xiaohuan and Leng, Yichong and Lv, Yuanjun and Zhao, Zhou and Zhou, Chang and Zhou, Jingren},
  booktitle={Proceedings of the 62nd Annual Meeting of the Association for Computational Linguistics},
  pages={1979--1998},
  year={2024},
  doi={10.18653/v1/2024.acl-long.109},
  url={https://aclanthology.org/2024.acl-long.109/}
}

@article{qin2025video,
  title={Video-XL-2: Towards Very Long-Video Understanding Through Task-Aware KV Sparsification},
  author={Qin, Minghao and Liu, Xiangrui and Liang, Zhengyang and Shu, Yan and Yuan, Huaying and Zhou, Juenjie and Xiao, Shitao and Zhao, Bo and Liu, Zheng},
  journal={arXiv preprint arXiv:2506.19225},
  year={2025}
}

@article{yan2025teleego,
  title={TeleEgo: Benchmarking Egocentric AI Assistants in the Wild},
  author={Yan, Jiaqi and Ren, Ruilong and Liu, Jingren and Xu, Shuning and Wang, Ling and Wang, Yiheng and Zhong, Xinlin and Wang, Yun and Zhang, Long and Chen, Xiangyu and others},
  journal={arXiv preprint arXiv:2510.23981},
  year={2025}
}

@inproceedings{long2025m3bench,
  title={Seeing, listening, remembering, and reasoning: A multimodal agent with long-term memory},
  author={Long, Lin and He, Yichen and Ye, Wentao and Pan, Yiyuan and Lin, Yuan and Li, Hang and Zhao, Junbo and Li, Wei},
  booktitle={International Conference on Learning Representations},
  year={2026},
  url={https://openreview.net/forum?id=PMz29A7Muq}
}

@inproceedings{wang2025videoitg,
  title={{VideoITG}: Multimodal Video Understanding with Instructed Temporal Grounding},
  author={Wang, Shihao and Chen, Guo and Huang, De-An and Li, Zhiqi and Li, Minghan and Liu, Guilin and Kautz, Jan and Alvarez, Jose M. and Zhang, Lei and Yu, Zhiding},
  booktitle={Proceedings of the IEEE/CVF Conference on Computer Vision and Pattern Recognition},
  pages={24640--24650},
  year={2026},
  url={https://openaccess.thecvf.com/content/CVPR2026/html/Wang_VideoITG_Multimodal_Video_Understanding_with_Instructed_Temporal_Grounding_CVPR_2026_paper.html}
}

@article{openai2026gpt5,
  title={OpenAI GPT-5 System Card},
  author={Singh, Aaditya and others},
  journal={arXiv preprint arXiv:2601.03267},
  year={2026},
  url={https://arxiv.org/abs/2601.03267}
}

@inproceedings{rege2026agentic,
  title={Agentic Very Long Video Understanding},
  author={Rege, Aniket and Sadhu, Arka and Li, Yuliang and Li, Kejie and Vinayak, Ramya Korlakai and Chai, Yuning and Lee, Yong Jae and Kim, Hyo Jin},
  booktitle={Proceedings of the 64th Annual Meeting of the Association for Computational Linguistics},
  pages={46575--46602},
  year={2026},
  doi={10.18653/v1/2026.acl-long.2161},
  url={https://aclanthology.org/2026.acl-long.2161/}
}

@article{jin2025videomem,
  title={VideoMem: Enhancing Ultra-Long Video Understanding via Adaptive Memory Management},
  author={Jin, Hongbo and Wang, Qingyuan and Zhang, Wenhao and Liu, Yang and Cheng, Sijie},
  journal={arXiv preprint arXiv:2512.04540},
  year={2025}
}

@misc{chen2026telemembuildinglongtermmultimodal,
      title={TeleMem: Building Long-Term and Multimodal Memory for Agentic AI}, 
      author={Chunliang Chen and Ming Guan and Xiao Lin and Jiaxu Li and Luxi Lin and Qiyi Wang and Xiangyu Chen and Jixiang Luo and Changzhi Sun and Dell Zhang and Xuelong Li},
      year={2026},
      eprint={2601.06037},
      archivePrefix={arXiv},
      primaryClass={cs.CL},
      url={https://arxiv.org/abs/2601.06037}, 
}

@misc{wang2025omemomnimemorypersonalized,
      title={O-Mem: Omni Memory System for Personalized, Long Horizon, Self-Evolving Agents}, 
      author={Piaohong Wang and Motong Tian and Jiaxian Li and Yuan Liang and Yuqing Wang and Qianben Chen and Tiannan Wang and Zhicong Lu and Jiawei Ma and Yuchen Eleanor Jiang and Wangchunshu Zhou},
      year={2025},
      eprint={2511.13593},
      archivePrefix={arXiv},
      primaryClass={cs.CL},
      url={https://arxiv.org/abs/2511.13593}, 
}

@misc{chen2026iterresearchrethinkinglonghorizonagents,
      title={IterResearch: Rethinking Long-Horizon Agents with Interaction Scaling}, 
      author={Guoxin Chen and Zile Qiao and Xuanzhong Chen and Donglei Yu and Haotian Xu and Wayne Xin Zhao and Ruihua Song and Wenbiao Yin and Huifeng Yin and Liwen Zhang and Kuan Li and Minpeng Liao and Yong Jiang and Pengjun Xie and Fei Huang and Jingren Zhou},
      year={2026},
      eprint={2511.07327},
      archivePrefix={arXiv},
      primaryClass={cs.AI},
      url={https://arxiv.org/abs/2511.07327}, 
}

@misc{yu2026agenticmemorylearningunified,
      title={Agentic Memory: Learning Unified Long-Term and Short-Term Memory Management for Large Language Model Agents}, 
      author={Yi Yu and Liuyi Yao and Yuexiang Xie and Qingquan Tan and Jiaqi Feng and Yaliang Li and Libing Wu},
      year={2026},
      eprint={2601.01885},
      archivePrefix={arXiv},
      primaryClass={cs.CL},
      url={https://arxiv.org/abs/2601.01885}, 
}

@inproceedings{
    zuo2025videolucy,
    title={VideoLucy: Deep Memory Backtracking for Long Video Understanding},
    author={Jialong Zuo and Yongtai Deng and Lingdong Kong and Jingkang Yang and Rui Jin and Yiwei Zhang and Nong Sang and Liang Pan and Ziwei Liu and Changxin Gao},
    booktitle={The Thirty-ninth Annual Conference on Neural Information Processing Systems},
    year={2025}
}

@misc{openai2026luna,
  title={{GPT-5.6 Luna}: Model Documentation},
  author={{OpenAI}},
  year={2026},
  url={https://developers.openai.com/api/docs/models/gpt-5.6-luna}
}

@misc{openai2026gpt61sol,
  title={{GPT-6.1 Sol}: Model Documentation},
  author={{OpenAI}},
  year={2026},
  url={https://developers.openai.com/api/docs/models/gpt-6.1-sol}
}

@misc{openai2026codex,
  title={{Codex}: Open-Source Coding Agent},
  author={{OpenAI}},
  year={2026},
  url={https://github.com/openai/codex},
  note={Accessed September 2026}
}

@article{gemmateam2026gemma4,
  title={{Gemma 4} Technical Report},
  author={{Gemma Team} and others},
  journal={arXiv preprint arXiv:2607.02770},
  year={2026},
  url={https://arxiv.org/abs/2607.02770}
}

@article{vteam2025glm,
  title={{GLM-4.5V} and {GLM-4.1V-Thinking}: Towards Versatile Multimodal Reasoning with Scalable Reinforcement Learning},
  author={{V Team} and Hong, Wenyi and Yu, Wenmeng and others},
  journal={arXiv preprint arXiv:2507.01006},
  year={2025},
  url={https://arxiv.org/abs/2507.01006}
}

@article{liu2026ministral3,
  title={{Ministral 3}},
  author={Liu, Alexander H. and Khandelwal, Kartik and Subramanian, Sandeep and others},
  journal={arXiv preprint arXiv:2601.08584},
  year={2026},
  url={https://arxiv.org/abs/2601.08584}
}

@article{nvidia2026nemotron3nanoomni,
  title={{Nemotron 3 Nano Omni}: Efficient and Open Multimodal Intelligence},
  author={{NVIDIA}},
  journal={arXiv preprint arXiv:2604.24954},
  year={2026},
  url={https://arxiv.org/abs/2604.24954}
}

@article{xu2025qwen3omni,
  title={{Qwen3-Omni} Technical Report},
  author={Xu, Jin and Guo, Zhifang and Hu, Hangrui and others},
  journal={arXiv preprint arXiv:2509.17765},
  year={2025},
  url={https://arxiv.org/abs/2509.17765}
}

@misc{qwen2026qwen35,
  title={{Qwen3.5}: Towards Native Multimodal Agents},
  author={{Qwen Team}},
  year={2026},
  url={https://qwen.ai/blog?id=qwen3.5}
}

@misc{qwen2026qwen38,
  title={{Qwen3.8-Max}: A New Bar for Coding and Cowork},
  author={{Qwen Team}},
  year={2026},
  url={https://qwen.ai/blog?id=qwen3.8}
}

@article{yan2026internvideo3,
  title={{InternVideo3}: Agentify Foundation Models with Multimodal Contextual Reasoning},
  author={Yan, Ziang and Xia, Sheng and Yu, Jiashuo and others},
  journal={arXiv preprint arXiv:2606.12195},
  year={2026},
  url={https://arxiv.org/abs/2606.12195}
}

@article{li2026videochat3,
  title={{VideoChat3}: Fully Open Video {MLLM} for Efficient and Generalist Video Understanding},
  author={Li, Xinhao and Zhu, Yuhan and Zeng, Xiangyu and others},
  journal={arXiv preprint arXiv:2607.14935},
  year={2026},
  url={https://arxiv.org/abs/2607.14935}
}

@article{liu2025videoxlpro,
  title={{Video-XL-Pro}: Reconstructive Token Compression for Extremely Long Video Understanding},
  author={Liu, Xiangrui and Shu, Yan and Liu, Zheng and Li, Ao and Tian, Yang and Zhao, Bo},
  journal={arXiv preprint arXiv:2503.18478},
  year={2025},
  url={https://arxiv.org/abs/2503.18478}
}

@inproceedings{liu2025longvideoagent,
  title={{LongVideoAgent}: Multi-Agent Reasoning with Long Videos},
  author={Liu, Runtao and Liu, Ziyi and Tang, Jiaqi and Ma, Yue and Pi, Renjie and Zhang, Jipeng and Chen, Qifeng},
  booktitle={Proceedings of the 64th Annual Meeting of the Association for Computational Linguistics},
  pages={40404--40416},
  year={2026},
  doi={10.18653/v1/2026.acl-long.1876},
  url={https://aclanthology.org/2026.acl-long.1876/}
}

@inproceedings{yeo2025worldmm,
  title={{WorldMM}: Dynamic Multimodal Memory Agent for Long Video Reasoning},
  author={Yeo, Woongyeong and Kim, Kangsan and Yoon, Jaehong and Hwang, Sung Ju},
  booktitle={Proceedings of the IEEE/CVF Conference on Computer Vision and Pattern Recognition},
  pages={25599--25609},
  year={2026},
  url={https://openaccess.thecvf.com/content/CVPR2026/html/Yeo_WorldMM_Dynamic_Multimodal_Memory_Agent_for_Long_Video_Reasoning_CVPR_2026_paper.html}
}

@inproceedings{jain2025sage,
  title={{SAGE}: Training Smart Any-Horizon Agents for Long Video Reasoning with Reinforcement Learning},
  author={Jain, Jitesh and Li, Jialuo and Ma, Zixian and Zhang, Jieyu and Kim, Chris Dongjoo and Lee, Sangho and Tripathi, Rohun and Gupta, Tanmay and Clark, Christopher and Shi, Humphrey},
  booktitle={Proceedings of the IEEE/CVF Conference on Computer Vision and Pattern Recognition},
  pages={41478--41488},
  year={2026},
  url={https://openaccess.thecvf.com/content/CVPR2026/html/Jain_SAGE_Training_Smart_Any-Horizon_Agents_for_Long_Video_Reasoning_with_CVPR_2026_paper.html}
}

@inproceedings{ding2025videozoomer,
  title={{VideoZoomer}: Reinforcement-Learned Temporal Focusing for Long Video Reasoning},
  author={Ding, Yang and Lai, Xin and Zhang, Yizhen and Li, Wei and Chu, Ruihang and Yang, Yujiu},
  booktitle={International Conference on Learning Representations},
  year={2026},
  url={https://proceedings.iclr.cc/paper_files/paper/2026/hash/221b2665260a582f7eb821b9871e8001-Abstract-Conference.html}
}

@article{yang2026paravt,
  title={{ParaVT}: Taming the Tool Prior Paradox for Parallel Tool Use in Agentic Video Reinforcement Learning},
  author={Yang, Zuhao and Zhang, Kaichen and Wang, Sudong and Wu, Keming and Yang, Zhongyu and Li, Bo and Qi, Xiaojuan and Lu, Shijian and Li, Xingxuan and Bing, Lidong},
  journal={arXiv preprint arXiv:2605.20342},
  year={2026},
  url={https://arxiv.org/abs/2605.20342}
}

@article{zou2026taskmem,
  title={Task-Focused Memorization for Multimodal Agents},
  author={Zou, Tao and He, Yichen and Qiu, Tian and Lin, Yuan and Li, Hang},
  journal={arXiv preprint arXiv:2605.31075},
  year={2026},
  url={https://arxiv.org/abs/2605.31075}
}

\clearpage
\appendix
\onecolumn

\makeatletter
\newcommand{\suppanchor}[2]{%
  \phantomsection\begingroup\def\@currentlabelname{#2}\label{#1}\endgroup}
\makeatother
\newcommand{\suppcontentsentry}[1]{%
  \hyperref[#1]{\ref*{#1}\ \nameref*{#1}}%
  \dotfill\hyperref[#1]{\pageref*{#1}}}
\newcommand{\suppsectionentry}[1]{%
  \textbf{\hyperref[#1]{\ref*{#1}.\ \nameref*{#1}}}%
  \dotfill\hyperref[#1]{\pageref*{#1}}}

\noindent\textbf{\Large Table of Contents}

\begin{itemize}[leftmargin=1em, label={}, itemsep=0.5em]
  \item \suppsectionentry{app:dataset}
    \begin{itemize}[leftmargin=1.5em, label={}, itemsep=0.2em]
      \item \suppcontentsentry{app:dataset:video}
      \item \suppcontentsentry{app:dataset:annotation}
      \item \suppcontentsentry{app:dataset:why-not-subsample}
      \item \suppcontentsentry{app:dataset:data-contamination}
        \begin{itemize}[leftmargin=1.5em, label={}, itemsep=0.2em, topsep=0.2em]
          \item \suppcontentsentry{app:dataset:web-search}
          \item \suppcontentsentry{app:dataset:year-scale}
        \end{itemize}
    \end{itemize}
  \item \suppsectionentry{app:exp}
    \begin{itemize}[leftmargin=1.5em, label={}, itemsep=0.2em]
      \item \suppcontentsentry{app:exp:prompts}
      \item \suppcontentsentry{app:exp:eval}
      \item \suppcontentsentry{app:exp:grounding-resolution}
    \end{itemize}
  \item \suppsectionentry{app:analysis}
  \item \suppsectionentry{app:limitations}
  \item \suppsectionentry{app:ethics}
  \item \suppsectionentry{app:compute}
\end{itemize}

\section{Benchmark Construction}
\label{app:dataset}

\subsection{Video}
\label{app:dataset:video}

Table~\ref{tab:clip-meta} presents the metadata for each video clip, including the start time, end time, and clip duration. 
For the Live Stream Subset, we report absolute UTC timestamps, as the original livestreams provide access to their true broadcast times. 
For the Gamer's Journey Subset, clips correspond to consecutive gameplay segments spanning multiple in-game chapters and are concatenated end-to-end. We therefore treat the entire sequence as continuous gameplay within a single day and report relative timestamps accordingly. 
For the Egocentric Life Subset, where the exact dates are unavailable, we report timestamps using relative day index combined with absolute time-of-day.

\begin{table}[!ht]
\centering
\caption{Temporal metadata of video clips across subsets. Each clip is annotated with its begin and end time, using relative timestamps or absolute UTC timestamps.}
\label{tab:clip-meta}
\footnotesize
\renewcommand{\arraystretch}{1.05}
\setlength{\tabcolsep}{1.6mm}

\begin{minipage}[t]{0.48\textwidth}
    \centering
    \caption{Video Timeline Accumulation of Gamer's Journey Subset.}
    \label{tab:video-timeline}
    \begin{tabular}{cllr}
    \toprule
    \textbf{\#} & \textbf{Begin} & \textbf{End} & \textbf{Dur.(s)} \\
    \midrule
    1 & Day 1 00:00:00 & Day 1 00:13:41 & 821 \\
    2 & Day 1 00:13:41 & Day 1 00:54:14 & 2,433 \\
    3 & Day 1 00:54:14 & Day 1 01:46:40 & 3,146 \\
    4 & Day 1 01:46:40 & Day 1 02:16:39 & 1,799 \\
    5 & Day 1 02:16:39 & Day 1 03:15:07 & 3,508 \\
    6 & Day 1 03:15:07 & Day 1 04:10:32 & 3,325 \\
    7 & Day 1 04:10:32 & Day 1 04:54:35 & 2,643 \\
    8 & Day 1 04:54:35 & Day 1 06:02:21 & 4,066 \\
    9 & Day 1 06:02:21 & Day 1 06:41:32 & 2,351 \\
    10 & Day 1 06:41:32 & Day 1 07:02:28 & 1,256 \\
    11 & Day 1 07:02:28 & Day 1 07:46:41 & 2,653 \\
    12 & Day 1 07:46:41 & Day 1 08:37:39 & 3,058 \\
    13 & Day 1 08:37:39 & Day 1 09:29:30 & 3,111 \\
    14 & Day 1 09:29:30 & Day 1 10:20:29 & 3,059 \\
    15 & Day 1 10:20:29 & Day 1 11:14:59 & 3,270 \\
    16 & Day 1 11:14:59 & Day 1 11:51:49 & 2,210 \\
    17 & Day 1 11:51:49 & Day 1 12:22:52 & 1,863 \\
    18 & Day 1 12:22:52 & Day 1 13:08:51 & 2,759 \\
    19 & Day 1 13:08:51 & Day 1 13:58:56 & 3,005 \\
    20 & Day 1 13:58:56 & Day 1 14:33:17 & 2,061 \\
    21 & Day 1 14:33:17 & Day 1 15:33:47 & 3,630 \\
    22 & Day 1 15:33:47 & Day 1 16:20:36 & 2,809 \\
    23 & Day 1 16:20:36 & Day 1 17:32:13 & 4,297 \\
    24 & Day 1 17:32:13 & Day 1 18:25:02 & 3,169 \\
    25 & Day 1 18:25:02 & Day 1 19:13:35 & 2,913 \\
    26 & Day 1 19:13:35 & Day 1 20:08:42 & 3,307 \\
    27 & Day 1 20:08:42 & Day 1 20:42:44 & 2,042 \\
    28 & Day 1 20:42:44 & Day 1 21:21:19 & 2,315 \\
    29 & Day 1 21:21:19 & Day 1 22:08:13 & 2,814 \\
    30 & Day 1 22:08:13 & Day 1 22:47:38 & 2,365 \\
    31 & Day 1 22:47:38 & Day 1 23:35:14 & 2,856 \\
    \bottomrule
    \end{tabular}
\end{minipage}%
\hfill
\begin{minipage}[t]{0.48\textwidth}
    \centering
    
    \caption{Video Timeline Accumulation of Live Stream Subset.}
    \label{tab:broadcast-meta}
    \resizebox{\linewidth}{!}{%
    \begin{tabular}{cllr}
    \toprule
    \textbf{\#} & \textbf{Begin time} & \textbf{End time} & \textbf{Dur.(s)} \\
    \midrule
    1 & 02-28T17:00:51Z & 03-01T04:59:01Z & 42,900 \\
    2 & 03-01T04:59:32Z & 03-01T06:52:59Z & 6,813 \\
    3 & 03-01T17:00:16Z & 03-01T21:10:17Z & 14,740 \\
    4 & 03-02T18:00:12Z & 03-02T22:00:22Z & 14,421 \\
    5 & 03-05T22:16:34Z & 03-05T23:13:35Z & 3,430 \\
    6 & 03-06T17:31:51Z & 03-06T21:00:34Z & 12,535 \\
    7 & 03-07T14:19:28Z & 03-07T16:11:38Z & 6,740 \\
    8 & 03-08T13:29:35Z & 03-08T15:02:49Z & 5,604 \\
    9 & 03-08T17:32:44Z & 03-08T18:12:22Z & 2,390 \\
    10 & 03-12T17:15:50Z & 03-12T20:06:57Z & 10,259 \\
    11 & 03-14T17:30:22Z & 03-14T20:46:29Z & 11,777 \\
    12 & 03-16T16:00:20Z & 03-16T20:14:46Z & 15,203 \\
    13 & 03-19T16:00:45Z & 03-19T18:55:28Z & 10,466 \\
    14 & 03-24T06:00:05Z & 03-24T12:19:50Z & 22,799 \\
    15 & 03-26T01:29:56Z & 03-26T07:39:35Z & 22,191 \\
    16 & 03-28T05:45:53Z & 03-28T10:35:38Z & 17,396 \\
    17 & 03-31T06:31:41Z & 03-31T11:54:51Z & 19,402 \\
    18 & 04-02T07:00:25Z & 04-02T13:31:41Z & 23,485 \\
    19 & 04-04T01:18:06Z & 04-04T10:57:34Z & 34,777 \\
    20 & 04-05T06:31:19Z & 04-05T12:14:24Z & 20,594 \\
    21 & 04-07T08:01:01Z & 04-07T14:05:28Z & 21,702 \\
    22 & 04-11T03:15:17Z & 04-11T08:25:16Z & 18,173 \\
    23 & 04-20T16:00:13Z & 04-20T22:30:30Z & 22,504 \\
    \bottomrule
    \end{tabular}
    }

    \vspace{0.6cm} 

    \caption{Video Timeline Accumulation of Egocentric Life Subset.}
    \label{tab:broadcast-meta-week1}
    \resizebox{\linewidth}{!}{%
    \begin{tabular}{cllr}
    \toprule
    \textbf{\#} & \textbf{Begin datetime} & \textbf{End datetime} & \textbf{Dur.(s)} \\
    \midrule
    1 & Day 1 11:09:42.08 & Day 1 22:05:49.11 & 39,367 \\
    2 & Day 2 10:44:25.06 & Day 2 22:58:25.00 & 44,040 \\
    3 & Day 3 11:17:27.02 & Day 3 22:51:33.07 & 41,646 \\
    4 & Day 4 10:48:20.00 & Day 4 22:24:34.18 & 41,774 \\
    5 & Day 5 11:00:31.00 & Day 5 23:29:46.08 & 44,955 \\
    6 & Day 6 09:49:33.00 & Day 6 22:16:59.01 & 44,846 \\
    7 & Day 7 11:56:08.17 & Day 7 20:16:15.08 & 30,007 \\
    \bottomrule
    \end{tabular}
    }
\end{minipage}
\end{table}

\FloatBarrier
\subsection{Annotation}
\label{app:dataset:annotation}

\textbf{Definition and Examples.} The definitions and number of QA types are listed in Table~\ref{tab:question-stats}, and examples are given in Table~\ref{tab:question-example}. Figures~\ref{fig:question-categroy} and~\ref{fig:video-category} show the distributions of question categories and video domains.

\begin{table}[!htb]
  \centering
  \scriptsize
  \setlength{\tabcolsep}{3pt}
  \caption{Statistics and definitions of question categories. The categories are sorted by the total number of samples.}
  \label{tab:question-stats}

  \begin{tabular*}{\linewidth}{@{\extracolsep{\fill}}l p{5.9cm} c c c c@{}}
    \toprule
    \textbf{Category} & \textbf{Definition} & \textbf{Day} & \textbf{Week} & \textbf{Month} & \textbf{Total} \\
    \midrule

    Counting &
    For the object(s) or event(s) mentioned, ask how many times they appear or repeat. &
    40 (20.00\%) & 59 (29.50\%) & 213 (23.96\%) & 312 (24.20\%) \\ \addlinespace

    Causal Reasoning &
    For an event mentioned, ask about its cause or the result it leads to. &
    20 (10.00\%) & 17 (8.50\%) & 151 (16.99\%) & 188 (14.58\%) \\ \addlinespace

    Entity Recognition &
    Identify a specific entity (object, person, or place) referenced in the question. &
    31 (15.50\%) & 23 (11.50\%) & 132 (14.85\%) & 186 (14.43\%) \\ \addlinespace

    Temporal Reasoning &
    Ask about temporal order, chronological sequencing, or duration of events. &
    31 (15.50\%) & 36 (18.00\%) & 85 (9.56\%) & 152 (11.79\%) \\ \addlinespace

    Hallucination Det. &
    Given a set of statements, ask which statements are correct or which are wrong. &
    31 (15.50\%) & 10 (5.00\%) & 71 (7.99\%) & 112 (8.69\%) \\ \addlinespace

    Event Recognition &
    Perform a recognition or identification of an event mentioned in the question. &
    11 (5.50\%) & 17 (8.50\%) & 80 (9.00\%) & 108 (8.38\%) \\ \addlinespace

    Lang. Content Recall &
    Ask about specific linguistic content (speech/ASR or on-screen text/OCR). &
    6 (3.00\%) & 15 (7.50\%) & 63 (7.09\%) & 84 (6.52\%) \\ \addlinespace

    Attribute Recognition &
    Ask about the attribute (e.g., color, appearance, size) of an object or person. &
    6 (3.00\%) & 8 (4.00\%) & 40 (4.50\%) & 54 (4.19\%) \\ \addlinespace

    Social Interaction &
    Ask about relationships, social roles, or the nature of interactions between people. &
    0 (0.00\%) & 9 (4.50\%) & 31 (3.49\%) & 40 (3.10\%) \\ \addlinespace

    State Change &
    For an object or scene mentioned, ask about how its state changes. &
    12 (6.00\%) & 0 (0.00\%) & 18 (2.02\%) & 30 (2.33\%) \\ \addlinespace

    Event Tracking &
    Ask when an event happened or when an object was acquired (retrospective tracing). &
    12 (6.00\%) & 6 (3.00\%) & 5 (0.56\%) & 23 (1.78\%) \\

    \bottomrule
  \end{tabular*}
\end{table}

\begin{table}[H]
\centering
\scriptsize
\caption{QA example of each category.}
\label{tab:question-example}
\setlength{\tabcolsep}{4pt}
\begin{tabular*}{\linewidth}{@{\extracolsep{\fill}}p{.22\linewidth}p{.74\linewidth}@{}}
\toprule
\textbf{Question sub-category}
&\textbf{QA pair example}  \\
\midrule

Counting   & Q: In the first live stream after returning from both the China trip and the Mongolia trip, how many times did [The Streamer] hear the song 'Sunshine Rainbow Little White Horse' while browsing videos on Discord?\\
   & A: 8 times\\
\midrule
Event Recognition   & Q: In the game FRAGPUNK played by [The Streamer], in the sixth round of the new match after the final score was 4:2, what event occurred that shocked [The Streamer]?\\
   & A: [The Streamer] and [The Streamer]'s teammates were all killed by an invisible enemy.\\
\midrule
Language Content Recall   & In Day 5, when the first-person protagonist arrives at the supermarket, what is written on the front of their tablet's case?\\
   & A: Scholar.\\
\midrule
State Change   & Q: What change occurred in the clothing of the character [The Streamer] was watching in the second video before hosting the talent show by [The Streamer]?\\
   & A: Red-Black\\
\midrule
Causal Reasoning   & Q: During [The Streamer]'s visit to Hong Kong, why did he still feel shocked after talking to a little boy upon leaving the gaming area and arriving at the parking lot?\\
   & A: The 12-year-old boy had an exceptionally mature voice.\\
\midrule
Event Tracking   & Q: During [The Streamer]'s trip to Mongolia, [The Streamer] was kissed on the cheek by a male fan in the car leaving the museum. When was the last time he was kissed by a male fan?\\
   & A: While walking on the street in Chengdu.\\
\midrule
Temporal Reasoning   & Q: In the event involving [The Streamer] and others linking microphones, what is the correct sequence? 1. A lady who changed many pairs of high heels and played football with them; 2. Three children, one of whom looks especially like Messi; 3. A chubby guy who challenged himself to eat a pizza within one minute.\\
   & A: 3, 2, 1\\
\midrule
Social Interaction   & Q: In [The Streamer]'s first live stream after returning from his trips to China and Mongolia, who appeared most frequently, even throughout the entire video, when he watched Coco's video about his China trip on Discord?\\
   & A: The man in the red floral shirt.\\
\midrule
Hallucination Detection   & Q: During [The Streamer]'s visit to China, which of the following statements are correct? 1. In a park in Chongqing, he played tennis for a while. 2. In a basketball court in Shanghai, he played basketball with Jackson Wang for a while. 3. In Yu Garden, Shanghai, he played a translation game with a Chinese guy, translating from English to Chinese. 4. At the end of his Shanghai visit, he greeted a guy with an injured ankle.\\
   & A: 1, 4\\

\midrule
Attribute Recognition& Q: After the player enters the Flaming Mountains chapter, what are the player's health and mana values, respectively, before the first challenge against Yinhu?\\
& A: 660, 360\\

\midrule
Entity Recognition& Q: On Day 1, what was the dessert made after dinner? \\
& A: Strawberry Cream Cupcake\\

\bottomrule
\end{tabular*}
\end{table}

\textbf{Annotation Cost.} Eight annotators were recruited to label the three subsets. Due to variations in temporal length, annotation costs differed across the data: the average time per sample was 19 minutes for the day-scale (Gamer's Journey), 33 minutes for the week-scale (Egocentric Life), and 51 minutes for the month-scale (Live Stream) subsets.

\subsection{Why Sub-sampling Cannot Substitute for Continuous Capture}
\label{app:dataset:why-not-subsample}
A natural question is whether the regime $T_{span} \gg T_{dur}$ at the day, week, or month scale can be obtained simply by sub-sampling existing long videos. It cannot. Sub-sampling can only enlarge the span up to the source's own physical duration. The longest commonly-released videos (movies, broadcasts, lectures) span only a few hours of physical time, so any sub-sampling protocol over them yields $T_{span}$ at most a few hours. Operationalizing $T_{span} \gg T_{dur}$ at the day, week, or month scale therefore requires \emph{continuous} physical-day-, week-, or month-scale capture of a single subject. Such data is not produced by post-hoc sub-sampling; it must be collected longitudinally.

\subsection{Data Contamination}
\label{app:dataset:data-contamination}
We investigated the risks of data contamination arising from the integration of search engine tools. Specifically, we analyzed year-scale data to assess the current state of contamination in large-scale evaluation.

\subsubsection{Web Search Impact}
\label{app:dataset:web-search}

We evaluated the impact of web search capabilities on a livestream subset by comparing the latest model from the Gemini family of multimodal models~\cite{gemini} in both offline and web-enabled configurations.

\paragraph{Prompting Strategy}
To facilitate efficient testing, we batched multiple queries into a single request, instructing the system to process them via a deep research workflow and return responses in a structured JSON format. Below is an example prompt containing 100 questions:

\begin{lstlisting}
[
    { "index": 1, "question": "{question 1}" },
    ...
    { "index": 100, "question": "{question 100}" }
]

Task: Thoroughly research and analyze these questions. Provide individual answers in the following JSON format:
[
    { "index": 1, "answer": "{answer 1}" },
    ...
    { "index": 100, "answer": "{answer 100}" }
]
\end{lstlisting}

\paragraph{Results and Analysis}
Without web access, this Gemini model achieved an accuracy score of 5.54. Enabling web search raised the score to 11.79, more than doubling accuracy through external information retrieval.

This gain also reflects the risk of dataset contamination. Since livestream content is often documented or discussed online, it becomes difficult to distinguish long-context reasoning from indirect exposure to ground-truth data. Evaluation of long-video understanding should therefore control external knowledge access and clearly separate \textit{closed-book} from \textit{open-book} settings.

\subsubsection{Year-scale Data Exploration}
\label{app:dataset:year-scale}

To explore year-scale data curation, we curated a dataset spanning the 19-year career of a professional athlete. This dataset comprises 140 match videos from the same competition, totaling approximately 582 GB and 236.42 hours of footage. 

We attempted to annotate 10 QA pairs, each requiring specific temporal verification. However, the annotation cost was nearly prohibitive: verifying a single QA pair took substantially longer than at the month scale. Even with semi-automatic annotation tools, human annotators were still required to perform exhaustive manual verification of clue intervals, yielding minimal efficiency gains. 

External metadata (news reports, match statistics) sped up annotation by letting annotators focus on specific segments, but it introduced contamination: frontier models such as GPT-5 and Gemini had already internalized this well-known information during pre-training and could answer a subset of questions from text-only knowledge. When equipped with web search, these models answered nearly all year-scale questions correctly, leaving little room to evaluate pure video-based reasoning.

As a result, we excluded the year-scale from the benchmark. These findings suggest that as external information becomes increasingly accessible, the evaluation of long-video understanding risks collapsing into a test of textual retrieval rather than visual reasoning. Future benchmarks should prioritize ``non-Googleable'' visual tasks to keep evaluation grounded in temporal-visual reasoning rather than parametric recall.

\section{Experiments}
\label{app:exp}

\FloatBarrier

\subsection{Evaluation Prompts}
\label{app:exp:prompts}

\begin{promptbox}{QA prompt for end-to-end MLLMs}
Answer the following question based on the video with a concise answer.

\noindent Question: \texttt{\textquotesingle\{HERE IS THE QUESTION\}\textquotesingle}
\end{promptbox}

\begin{promptbox}{Grounding prompt for video agents and video-LLMs}
Find time intervals in the video when the query occurs.

\noindent Query: \texttt{\textquotesingle\{HERE IS THE QUESTION\}\textquotesingle}

\noindent Provide all possible intervals in seconds. Format for each interval:
\texttt{\textquotesingle xx.xx seconds - xx.xx seconds\textquotesingle}. Multiple intervals are linked by
\texttt{\textquotesingle\ and\ \textquotesingle}. Output the intervals only, do not output anything else.
\end{promptbox}

\subsection{Detailed Evaluation Protocols}
\label{app:exp:eval}

\suppanchor{app:exp:ref-code}{Ref@N evaluation code}
\textbf{Ref@N evaluation code.}
We provide Python code to provide a better understanding of the evaluation of reference grounding.

\begin{lstlisting}
from typing import List, Tuple, Set

Interval = Tuple[float, float]

def Ref_N(
    intervals_a: List[Interval],
    intervals_b: List[Interval],
    total_seconds: float,
    bucket_size: float = 300.0,
) -> float:
    def intervals_to_buckets(intervals: List[Interval]) -> Set[int]:
        buckets: Set[int] = set()
        for s, e in intervals:
            # clamp
            s = max(0.0, s)
            e = min(total_seconds, e)
            if s >= e:
                continue

            start = int(s // bucket_size)
            end = int((e - 1e-9) // bucket_size)
            buckets.update(range(start, end + 1))
        return buckets

    buckets_a = intervals_to_buckets(intervals_a)
    buckets_b = intervals_to_buckets(intervals_b)

    if not buckets_a and not buckets_b:
        return 0.0

    return len(buckets_a & buckets_b) / len(buckets_a | buckets_b)
\end{lstlisting}

\Needspace{4\baselineskip}
\suppanchor{app:exp:rubric-grading}{Rubric-based grading}
\textbf{Rubric-based grading.}
We evaluate each model prediction against the frozen rubric for its question, using GPT-5.6-luna as the rubric-based verifier. The grading inputs are the question, reference answer, rubric, and prediction. The rubric specifies the required facts, accepted expressions and aliases, incorrect claims, and conditions for assigning each score. The verifier assesses the prediction against these criteria and provides an integer score from 0 to 5 with a brief justification. Video evidence is used to establish the rubric during construction; grading uses the resulting textual criteria.

We aggregate the per-question scores as
\[
100\times\frac{\sum_i s_i}{5N},
\]
where $N$ is the full split size. Missing or unusable answers receive zero. For auxiliary strict-accuracy analyses, only scores of 4 or 5 count as correct.

\Needspace{8\baselineskip}
\suppanchor{app:exp:verifier-prompt}{Verifier prompt}
\textbf{Verifier prompt.}
For readability, we present the grading instructions in single-prediction form below. The reference answer states the target response, while the rubric defines acceptable alternatives and the conditions for full or partial credit. The prediction is the model response being assessed. The output contains its score and a short explanation tied to the rubric. The worked example on the next page illustrates the distinction between an omitted detail and an incorrect claim.

\begin{promptbox}{Verifier system prompt}
Evaluate one model prediction using the supplied question, reference answer,
and frozen question-specific rubric.

Check which required facts the prediction provides, which it omits, and which
it contradicts. Interpret equivalent expressions using the accepted phrasings
and aliases in the rubric.

Apply the rubric's scoring conditions, including any requirements for exact
values, temporal tolerances, set membership, or item order. Distinguish omissions
from incorrect claims as specified in the rubric. Do not revise the rubric,
introduce additional requirements, or use other models' answers as evidence.

Assign one integer score from 0 to 5 according to the question-specific rules.
Briefly explain which rubric conditions support that score.

Return exactly one JSON object with two fields:
\begin{quote}
\texttt{"score"}: an integer from 0 through 5;\\
\texttt{"reason"}: a brief rubric-based justification.
\end{quote}
Do not include markdown or any text outside the JSON object.
\end{promptbox}

\begin{promptbox}{Verifier user-message template}
\noindent\textbf{QUESTION:}\\
\texttt{\{question\}}

\noindent\textbf{REFERENCE ANSWER:}\\
\texttt{\{reference\_answer\}}

\noindent\textbf{FROZEN RUBRIC:}\\
\texttt{\{frozen\_rubric\_json\}}

\noindent\textbf{MODEL PREDICTION:}\\
\texttt{\{model\_prediction\}}

\noindent Evaluate this prediction according to the supplied rubric.
\end{promptbox}

The four placeholders are filled with the question, reference answer, complete frozen rubric, and one model prediction. The returned \texttt{score} is used in the benchmark aggregate, while \texttt{reason} explains the assessment.

\clearpage
\begin{outerbox}[title={Worked example: interpreting a location rubric}]
\suppanchor{app:exp:rubric-example}{Worked example: interpreting a location rubric}
\small
This example shows three recorded predictions for the same question and their GPT-5.6-luna assessments. The relevant rubric clauses are reproduced below; predictions, scores, and justifications are quoted from the evaluation records.

\begin{chatbox}{Question and reference answer}
\textbf{Question:} In Day 7, where was the second camera that the first-person protagonist dismantled located?\\
\textbf{Canonical answer:} Above the projector
\end{chatbox}

\begin{chatbox}{Relevant fields of the frozen rubric}
\textbf{Required fact:} The second camera dismantled on Day 7 was located above the projector.\\
\textbf{Accepted expressions:} ``above the projector''; ``over the projector''; ``mounted directly above the projector''; ``on the wall above the projector''; ``high on the wall, directly over the projector''.

\smallskip
\textbf{5:} States that the camera was above, directly over, or mounted above the projector.\\
\textbf{3:} States a compatible broad location such as on the wall or ceiling, but omits the essential projector relation without giving a conflicting landmark.\\
\textbf{2:} Gives an elevated or wall-area location but substitutes an important incorrect landmark, such as a doorway, computer area, or monitor room.\\
\textbf{0:} Gives a contradictory location such as a kitchen, bedroom, bathroom, bed, desk, or table.
\end{chatbox}

\begin{chatbox}{Prediction 1 and assessment}
\textbf{Answer:} The second camera was located on the wall near the ceiling.\\
\textbf{Score:} 3/5.\\
\textbf{Grader justification:} Wall near the ceiling is a compatible broad elevated location but omits the projector relation.
\end{chatbox}

\begin{chatbox}{Prediction 2 and assessment}
\textbf{Answer:} The second camera was located on the wall near the door.\\
\textbf{Score:} 2/5.\\
\textbf{Grader justification:} Gives a wall location but incorrectly identifies the door as the landmark.
\end{chatbox}

\begin{chatbox}{Prediction 3 and assessment}
\textbf{Answer:} in the bathroom\\
\textbf{Score:} 0/5.\\
\textbf{Grader justification:} Contradictory bathroom location.
\end{chatbox}

The first prediction leaves the reference landmark unspecified, whereas the second supplies a wrong landmark. The rubric assigns different partial credit to these two errors. The third prediction gives a contradictory room location and receives zero. Each score follows the same rubric, rather than the prediction's wording alone.
\end{outerbox}

\FloatBarrier

\clearpage
\subsection{Effect of Temporal Bin Width}
\label{app:exp:grounding-resolution}

\begin{figure}[htbp]
  \centering
  \includegraphics[width=\linewidth]{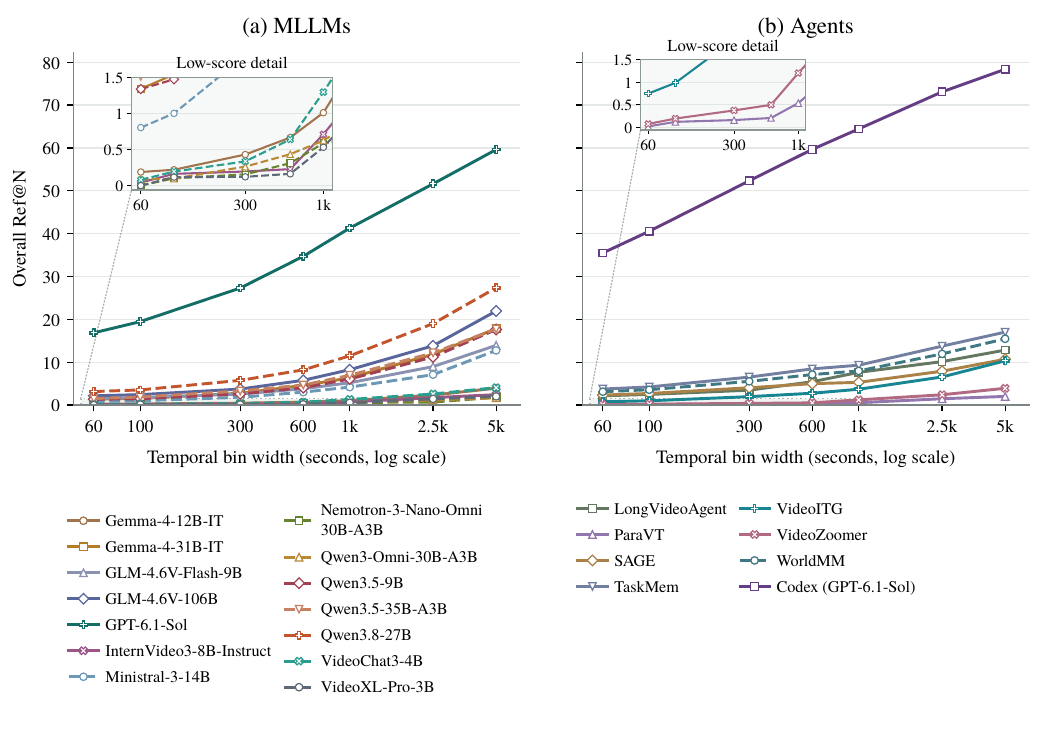}
  \caption{Overall Ref@N across temporal bin widths for MLLMs and agents. Insets enlarge the shaded low-score regions. The Codex (GPT-6.1-Sol) curve uses the timestamps cited in its answer runs. Predictions are held fixed as bin width varies.}
  \label{fig:grounding-resolution}
\end{figure}

We vary the temporal bin width to examine how grounding performance depends on localization tolerance. Figure~\ref{fig:grounding-resolution} compares 22 methods at 60, 100, 300, 600, 1000, 2500, and 5000 seconds.

For end-to-end GPT-6.1-Sol, widening the bin from 60 to 5000 seconds raises Ref@N from 16.84 to 59.60. Over the same range, Codex (GPT-6.1-Sol) rises from 35.48 to 78.34. A 5000-second bin spans more than an hour, so predictions and reference clues can match even when they point to different moments.

The answer-score sweep in Figure~\ref{fig:overall}(b) also includes InternVL3.5-38B, MiniCPM-o-4.5-9B, Molmo2-8B, and LongVILA-R1-7B. Their archived 256K runs yielded no usable temporal intervals. VideoChat3-4B and VideoXL-Pro-3B are reported at their model-specific settings in Table~\ref{tab:main_results}.

\FloatBarrier
\clearpage
\section{Qualitative Examples}
\label{app:analysis}

Supplementary Qwen3.8-27B thinking runs use 250 sampled frames and a 256K context limit. Each is paired with the archived Codex (GPT-6.1-Sol) trajectory for the same question. We summarize how Codex searched the recordings and inspected candidate events. Cited evidence and annotations use the benchmark playback timeline. The rubrics below state the effective scoring rules, including deterministic rules for structured answers. Qwen thinking is reproduced in full with names and account identifiers anonymized; the globe emoji is rendered as [globe symbol].

\begin{outerbox}[title={Example 1: purchase order at a snack shop}]
\suppanchor{app:analysis:example-1}{Example 1: purchase order at a snack shop}
\begin{chatbox}{Question}
During the streamer's visit to Hong Kong, what foods did the streamer purchase one after another at the snack shop? 1. Sweet soup; 2. Beef offal; 3. Fried potatoes?
\end{chatbox}

\begin{chatbox}{Reference answer and frozen rubric}
\textbf{Answer:} Beef offal, fried potatoes, then sweet soup: 2, 3, 1.\\
\textbf{Annotated evidence:} [290859, 291592] seconds.

\smallskip
\textbf{Accepted forms:} Option numbers or unambiguous food names in the predicted order; adjacent duplicate restatements count once.\\
\textbf{5:} The exact complete sequence 2, 3, 1.\\
\textbf{Partial credit:} Divide the longest matching in-order subsequence length by the longer of the predicted and reference sequence lengths, multiply by five, round halves upward, and cap at 3.\\
\textbf{0:} No matching item, an empty answer, or an unparseable sequence. The deterministic rule does not assign 4; incorrect orders cannot receive full credit.
\end{chatbox}

\end{outerbox}
\begin{chatbox}[breakable, title after break={Example 1: Qwen3.8-27B thinking (continued)}, before skip=6pt, after skip=6pt, before upper={\setlength{\parskip}{3pt}}]{Example 1: Qwen3.8-27B complete thinking}
The user is asking about a specific segment of the stream where the streamer visits a snack shop in Hong Kong and buys food one after another. The options are: 1. Sweet soup; 2. Beef offal; 3. Fried potatoes.

Looking at the frames, the Hong Kong segment appears to be roughly from frame 199 (262944.5 seconds) to frame 222 (317910 seconds). Let me look at the frames in this range.

Frame 199 (262944.5): The streamer is outside a LOG-ON building, wearing a red ``streamer 7'' jersey.\newline
Frame 200 (265915.5): The streamer is walking near a hedge.\newline
Frame 201 (268886.5): The streamer is talking with staff near an ice sculpture.\newline
Frame 202 (271857.5): The streamer is walking with a crowd.\newline
Frame 203 (273343.5): The streamer is on a soccer court.\newline
Frame 204 (274829.0): Inside a car.\newline
Frame 205 (277800.0): Inside a car, the driver is on the phone.\newline
Frame 206 (280771.0): Looking at a smartphone in the car.\newline
Frame 207 (283742.0): On a street with a Starbucks, the streamer is leaning out of a bus window.\newline
Frame 208 (284485.5): Inside the bus, holding a phone and a checkered bag.\newline
Frame 209 (286713.5): Reclining in the car.\newline
Frame 210 (289684.5): Looking at a phone in the car.\newline
Frame 211 (292655.5): The streamer is holding a plastic bag containing food (looks like fried food, probably fried potatoes or something), and a man in a floral jacket is holding a phone. The streamer is holding a bag of what appears to be fried food.\newline
Frame 212 (294141.0): On a boat with people.\newline
Frame 213 (296369.5): Sitting on the boat reading a paper.

Hmm, the question is asking about buying food one after another at a snack shop. The sampled frames don't clearly show a snack shop scene with sweet soup, beef offal, and fried potatoes. The frames are sampled at roughly 1500-1500 second intervals, so the snack shop scene may have been missed between samples.

Wait, but the question expects an answer based on the provided frames. Let me check the frames again. The question: ``During the streamer's visit to Hong Kong, what food did the streamer buy one after another at the snack shop? 1. Sweet soup; 2. Beef offal; 3. Fried potatoes?''

This looks like a multiple-choice question where the answer is probably a combination like ``2 and 3'' or a specific one. Since I only have the sampled frames, I need to find evidence. Frame 211 shows the streamer holding a plastic bag of food that looks like fried food (probably fried potatoes or fried dough). The man in the floral jacket is holding a phone, probably paying.

Actually, this is a known stream: streamer's China trip. In the Hong Kong segment, there was a scene where he ate at a street food stall and bought things like ``fried potatoes'' (potato chips?) or ``beef offal'' (beef tripe)? Hmm.

Given the options, the answer is probably ``2 and 3'' or just one. Let me think about what actually happened in the stream. streamer's China trip (April 2025) included Beijing (Forbidden City, Shaolin temple), Chongqing, Hong Kong, and Mongolia. In Hong Kong, there was a scene where he went to a snack shop and bought sweet soup? Hmm.

Actually, I recall a scene where streamer bought ``fried potatoes'' (chips) and ``beef offal''? Not sure.

The question format: ``what food did the streamer buy one after another at the snack shop? 1. Sweet soup; 2. Beef offal; 3. Fried potatoes?'' Probably the answer is ``2 and 3'' (beef offal and fried potatoes). Or the answer might be a list: ``beef offal and fried potatoes.''

Since I have to answer based only on the frames, and the frames don't clearly show the purchase, maybe the intended answer is based on frame 211 where he's holding a bag of fried food. Hmm.

Let me look at frame 211 more carefully: The streamer in a red jersey is holding a plastic bag containing brownish fried food (looks like fried potatoes or fried dough). The man in the floral jacket is holding a phone (probably paying). The crowd is around. This could be at a snack shop where he bought fried potatoes.

But ``one after another'' suggests multiple purchases. The options are numbered 1, 2, 3. The answer is probably ``2 and 3'' meaning beef offal and fried potatoes.

Given the ambiguity, the safest answer is: ``Beef offal and fried potatoes (options 2 and 3).''

Evidence: The Hong Kong segment is roughly 262944.5 to 317910. The snack shop purchase is probably around frame 211 (292655.5). Let me give evidence intervals around there: [292655.5, 294141.0]? But that interval includes the boat scene. Hmm.

Actually, the purchase scene might be between frames 210 and 212. Frame 211 shows him holding the food bag. So the evidence is [291170, 294141]? I need to use the frame timestamps as boundaries. The evidence intervals must be absolute timestamps; I can use the frame sample times as boundaries.

Let me think about the actual stream content: In Hong Kong, streamer visited a ``snack shop'' and bought ``sweet soup''? Hmm, I recall a scene where he drank ``sweet soup'' (like a dessert soup) and ate ``beef offal''? Not sure.

Let me think about the known video: ``streamer in Hong Kong'' - he visited the Victoria Harbour cruise, and there was a street food scene where he ate ``fried potatoes'' (potato chips?) Actually, I recall a scene where he bought ``beef offal'' (beef tripe) from a street vendor? Hmm.

Alternatively, the question might be from a quiz where the correct answer is ``2. Beef offal; 3. Fried potatoes'' meaning he bought both. The phrase ``one after another'' suggests a sequence.

Given the constraints, I'll answer: ``He bought beef offal and fried potatoes (options 2 and 3).''

Evidence: Frame 211 at 292655.5 shows him holding a bag of fried food after the purchase. Also frame 208 at 284485.5 shows him holding a checkered bag? That's a bag, not food.

Hmm, but maybe the snack shop scene is actually between frames 207 and 211. Frame 207 shows a street with a Starbucks and a bus. Frame 211 shows the purchase. So the evidence is [283742.0, 292655.5]? That's a wide range.

The instructions say: ``The evidence field must contain up to 8 absolute video time intervals supporting the answer.'' I can provide intervals covering the relevant segment. Since I can't pinpoint exactly, I'll provide [292655.5, 294141.0] as the purchase moment? But the end must be greater than the start; that's fine.

Actually, maybe I should reconsider: The question might be a trick: the answer is ``2 and 3'' because sweet soup wasn't bought. Or the answer is ``1 and 2''? Let me think about what's plausible: At a Hong Kong snack shop, typical items: sweet soup (like a dessert), beef offal (like a tripe soup), fried potatoes (chips). The streamer buying ``one after another'' - I recall a scene where streamer bought ``fried potatoes'' and then ``beef offal''? Hmm.

Let me try to recall the actual stream: streamer's China trip Part 2 (Hong Kong). There's a scene where he goes to a food stall and the vendor gives him ``fried potatoes'' (like potato chips) and he eats them. Then he buys ``beef offal''? I really can't remember.

Given the question is likely from a benchmark with a correct answer, the expected answer is probably ``2 and 3'' or ``beef offal and fried potatoes.'' I'll go with that.

Alternatively, the expected answer might be a single number like ``3.'' The question lists three options; ``what food did the streamer buy one after another'' - the answer might be ``2 and 3.''

I'll answer: ``Beef offal and fried potatoes (2 and 3).''

Evidence: Use the interval around frame 211: [292655.5, 294141.0]. Maybe also [291170.0, 292655.5]? I only have the sample times; I can create intervals between consecutive sample times. Frame 210 is 289684.5, frame 211 is 292655.5, frame 212 is 294141.0. The purchase is probably just before frame 211. So the interval [289684.5, 292655.5] covers the approach, and [292655.5, 294141.0] covers holding the food. I'll provide both.

Hmm, but the evidence should support the answer. Since the frames don't explicitly show beef offal, the evidence is weak. But the task says to answer using only the sampled frames; if I can't localize, use an empty list. But I should try.

Given the uncertainty, maybe the intended correct answer is ``2 and 3'' with evidence around 292655.5. I'll include [292655.5, 294141.0].

Let me double-check frame 211: The streamer is holding a transparent plastic bag containing brownish fried items - could be fried potatoes or fried dough. The man in the floral jacket is holding a phone, probably paying. The crowd is around. This matches ``buying food at a snack shop.''

So the answer: ``He bought beef offal and fried potatoes (options 2 and 3).''

I'll format the JSON accordingly.

Actually, wait. Let me reconsider the possibility that the answer is ``1 and 3'' or something. The phrase ``one after another'' might refer to him buying sweet soup first, then fried potatoes? Hmm.

Let me think about the actual video content from memory of streamer's Hong Kong vlog: I recall a scene where he was at a street food stall and bought ``fried potatoes'' (chips) and ate them, and also ``beef offal'' (like a tripe skewer)? I really can't remember.

Alternatively, this question might be from a Chinese quiz dataset about this stream, and the correct answer is ``2 and 3.'' I'll commit to that.

Final answer JSON: \{``answer'': ``Beef offal and fried potatoes (2 and 3)'', ``evidence'': [[292655.5, 294141.0]]\}

Maybe also include [289684.5, 292655.5]. I'll include two intervals.

Let me make sure the boundaries are decimals: 289684.5, 292655.5, 294141.0. Good.
\end{chatbox}

\noindent\begin{minipage}{\linewidth}
\begin{chatbox}{Example 1: Qwen3.8-27B prediction and score}
\textbf{Final answer:} He bought the beef offal and the fried potatoes (options 2 and 3), not the sweet soup.\\
\textbf{Rubric score:} 3/5. Two foods retain their reference order, but sweet soup is omitted and denied. The longest matching subsequence has length 2 of 3; rounding and the partial-credit cap give 3.
\end{chatbox}

\begin{chatbox}{Example 1: Codex (GPT-6.1-Sol) tool-use trajectory}
\textbf{Search.} Codex sampled one frame every five minutes from the April 4 Hong Kong recording to locate food stops.\\
\textbf{Closer inspection.} It sampled the candidate food-stop segment every ten seconds, then examined individual frames of the counter and fried snacks.

\par\smallskip
\textbf{Final answer:} \emph{1 and 2: sweet soup and beef offal. The observed order is beef offal before sweet soup.}\\
\textbf{Cited evidence:} Beef offal at 290011--290081 and sweet soup at 290371--290421.\\
\textbf{Rubric score:} 3/5. The two foods are in the correct relative order. Fried potatoes are missing.\\
\textbf{Evidence check:} Both cited intervals precede the annotated purchase interval.
\end{chatbox}
\end{minipage}

\clearpage
\begin{outerbox}[title={Example 2: left--right choices in Split or Steal}]
\suppanchor{app:analysis:example-2}{Example 2: left--right choices in Split or Steal}
\begin{chatbox}{Question}
In the 'Split or Steal' game organized by the streamer, what were the final choices made by the two girls on the left and right in the fourth round of the game?
\end{chatbox}

\begin{chatbox}{Reference answer and frozen rubric}
\textbf{Answer:} The girl on the left chose to split, and the girl on the right chose to steal.\\
\textbf{Annotated evidence:} [153393, 153567] seconds.

\smallskip
\textbf{5:} Both choices with the correct assignment: left = Split, right = Steal. An unambiguous left-to-right list is accepted.\\
\textbf{4:} Correct assignment with a minor omission, such as not repeating the round number.\\
\textbf{3:} One correct choice, or both labels without a clear side assignment.\\
\textbf{2:} Reversed assignment, or a correct choice mixed with an important side/entity error.\\
\textbf{1:} Topical overlap without the requested assignment; the frozen rubric places 'both Steal' here.\\
\textbf{0:} Empty, unrelated, nonsensical, refusal, or an unsupported claim that the event is absent.
\end{chatbox}

\end{outerbox}
\begin{chatbox}[breakable, title after break={Example 2: Qwen3.8-27B thinking (continued)}, before skip=6pt, after skip=6pt, before upper={\setlength{\parskip}{3pt}}]{Example 2: Qwen3.8-27B complete thinking}
The user is asking about the ``Split or Steal'' game that the streamer organized. Looking at the frames, frames 57 and 59 show a poll with the question ``who moves on to the final round?'' with options ``poll participant 1'' and ``poll participant 2''. Hmm, this isn't Split or Steal.

Let me look at the frames more carefully. Frame 57 (75020.5 seconds) shows a YouTube poll ``who moves on to the final round?'' with options ``poll participant 1'' and ``poll participant 2''. Frame 59 (77249.0 seconds) shows a poll ``do we keep poll participant 4 for final round'' with Yes 64\%, No 36\%.

Hmm, the question mentions ``the two girls on the left and right'' in the fourth round of the ``Split or Steal'' game. Wait, maybe the question is referring to a different game. Let me scan the frames again.

Actually, the question might be about a game where two girls (participants) make choices. Frame 57 shows a poll with two options. But ``the two girls on the left and right''—in frame 57, the layout shows four video panels: top-left a guy holding a phone, top-right the streamer in a red shirt, bottom-left the streamer in a red shirt, bottom-right a guy in a white shirt. Hmm.

Wait, maybe the question is about the ``Split or Steal'' game that the streamer organized with two participants, and the fourth round choices were ``split'' and ``steal''? The question asks: ``What were the final choices made by the two girls on the left and right in the fourth round of the game?''

Looking at the frames, I don't see any explicit ``Split or Steal'' game. But maybe the poll in frame 57 is part of a game where two girls (poll participant 1 and poll participant 2?) vote. Hmm.

Let me think about the context: this is streamer's stream. There was a famous stream where he played ``Split or Steal'' with two girls? Actually, I recall streamer did a ``split or steal'' game with two people? Hmm.

Or, the question might be a trick: the answer might be that the left girl chose ``split'' and the right girl chose ``steal,'' or both chose something. But I need evidence from the frames.

Let me carefully look at the frames around 75020-77249 seconds. Frame 57: poll ``who moves on to the final round?'' options: ``poll participant 1'' 55\%, ``poll participant 2'' 4\%, ``poll participant 3'' 13\%. Hmm, these look like usernames of participants in some game.

Frame 59: poll ``do we keep poll participant 4 for final round'' Yes 64\%, No 36\%.

Hmm, this doesn't match ``Split or Steal.''

Maybe the question is referring to a different part: frames 47-50 show a ``a footballer fan'' game and a Kahoot ``Name the logo.'' Frames 50-53 show a rebus puzzle game. Frames 54-56 show video calls.

Actually, wait. The question might be about a game where the streamer organized ``Split or Steal'' with two girls connected via video call, and in the fourth round, the left girl chose ``split'' and the right girl chose ``steal.'' But the frames don't show that text.

Let me re-read the question: ``In the 'Split or Steal' game organized by the streamer, what were the final choices made by the two girls on the left and right in the fourth round of the game?''

Hmm, maybe the answer is: the girl on the left chose ``split'' and the girl on the right chose ``steal.'' But I need to verify with the frames.

Looking at frame 57 again: the poll window shows ``Top chat'' and the poll ``who moves on to the final round?'' with options ``poll participant 1'' 55\%, ``poll participant 2'' 4\%, ``poll participant 3'' 13\%. This is about who moves on to the final round, not split/steal.

Hmm, maybe the question is mislabeled and actually refers to the poll where the two girls (participants) voted? The phrase ``the two girls on the left and right'' might refer to the two video call participants on the left and right of the streamer.

Actually, wait. Let me think about the actual streamer stream: there was a stream where he played ``Split or Steal'' with two girls? I don't recall. But there is a known clip: streamer ``split or steal'' with two girls? Hmm.

Alternatively, the question might be about the ``final round'' poll: the two girls' final choices were ``poll participant 1'' and ``poll participant 2''? That doesn't make sense.

Let me consider that the question might be a hallucination test: the correct answer might be that the frames don't show a ``Split or Steal'' game, so the answer is ``cannot be determined.'' But the instructions say to answer using only the sampled frames, and to return an empty list if evidence can't be localized. But the question expects a specific answer.

Hmm, but maybe the ``Split or Steal'' game is the one in frames 57 and 59: the poll ``who moves on to the final round?'' is actually the ``steal'' mechanism? Hmm.

Let me look at the timestamps: frame 57 at 75020.5 seconds shows the poll with ``poll participant 1'' leading at 55\%. Frame 59 at 77249.0 seconds shows a different poll ``do we keep poll participant 4 for final round'' Yes 64\%.

The question says ``the fourth round of the game.'' The poll might be for the fourth round. The two girls on the left and right: in frame 57, the bottom-left panel shows the streamer, the bottom-right shows a guy in a white shirt; top-left shows a guy in a black hoodie holding a phone; top-right shows the streamer in a red shirt. No girls.

Hmm, maybe the question is about a completely different segment: frames 40-44 show an Omegle-style ``Country [globe symbol] I am'' game with a spinning wheel. Not split or steal.

Frames 54-56: video call with a guy in a red beanie, an Asian guy, a guy with a guitar. Not girls.

Frames 58-60: a guy in a blue shirt with a timer. Not.

So where are the two girls? Maybe the question is fabricated to test whether the model will hallucinate. The correct response might be that the video doesn't contain a ``Split or Steal'' game with two girls, so the answer is that it cannot be determined from the frames. But the schema requires a concise final answer and evidence intervals.

Given the instruction ``answer using only the sampled video frames,'' if the event isn't present, we should say so and provide an empty evidence list.

But wait, maybe I'm missing something: the poll in frame 57: ``who moves on to the final round?'' with options ``poll participant 1'' and ``poll participant 2''. The names in parentheses might be descriptions of the girls' choices? ``poll participant 1 alias''? No.

Actually, ``poll participant 1'' might be a girl's username, and ``poll participant 2'' might be another person. The poll is deciding who moves on to the final round. This might be part of a ``split or steal'' game where each round the chat votes? Hmm.

Let me think about the actual stream: streamer did a ``streamer vs streamer'' game? There was a ``split or steal'' with two streamers? Hmm.

Alternatively, the question might refer to a game where two girls (the streamer's friends?) are on the left and right of the screen and each round they choose split or steal. The fourth round's final choices: left chose ``split,'' right chose ``steal.'' Without audio, we can't know.

Given the ambiguity, the safest answer is: the frames don't show a ``Split or Steal'' game; the closest is the poll in frames 57/59 about who moves on to the final round. But the question explicitly asks about the two girls' final choices in the fourth round.

Hmm, maybe the intended answer is: the girl on the left chose ``split'' and the girl on the right chose ``steal.'' Many quiz questions about videos like this are generated from actual video content with captions. The video might have a segment where the streamer plays ``split or steal'' with two girls via video call, and in the final (fourth) round, the left girl chose ``split'' and the right girl chose ``steal,'' resulting in... The frames at 75020.5 and 77249 show polls, which might be the chat voting on the outcome.

Actually, wait: the poll ``who moves on to the final round?'' with options ``poll participant 1'' 55\% and ``poll participant 2'' 4\% and ``poll participant 3'' 13\%. Total 72\%. The third option ``poll participant 3'' might be a third participant. This looks like a voting game among three people? Not two girls.

I think the question is unanswerable from the frames; the expected answer might be ``The girl on the left chose split and the girl on the right chose steal'' but I can't verify.

Let me consider the possibility that the question is a trick: the correct answer is that no such game appears in the sampled frames, so the answer: ``The frames do not show a 'Split or Steal' game; therefore the choices cannot be determined.'' Evidence: empty list.

Given the instruction ``return exactly one JSON object'' and ``use an empty list if evidence cannot be localized,'' I think the intended behavior for a question about content not present is to state it's not shown. But the question seems to assume the game exists. Hmm.

Let me double-check all frames for any ``split'' or ``steal'' text. Frame 47: ``Only a true a footballer fan can conquer this game'' with Quit/Continue. Frame 48: Kahoot ``Name the logo'' with McDonald's etc. Frame 50: rebus puzzle round 5. Frame 57: poll. Frame 59: poll. No ``split or steal.''

The two girls: frame 30 shows a girl (blonde) and a boy and a woman with glasses in a game cutscene (that's the game's story, not the streamer's game). Frame 26 shows a woman in a game cutscene. Not the streamer's game.

Thus, the ``Split or Steal'' game organized by the streamer is not present in the sampled frames. Therefore the answer: cannot be determined / not shown.

But maybe the question refers to the poll in frame 57 as the ``fourth round'' of some game, and the ``two girls on the left and right'' are the participants whose names are ``poll participant 1'' and ``poll participant 2''? The final choices: poll participant 1 (55\%) and poll participant 2 (4\%)? That doesn't map to split/steal.

I think the best answer is: the sampled frames don't contain a ``Split or Steal'' game with two girls, so the choices cannot be identified. Provide an empty evidence list.

But the question demands ``what were the final choices made by the two girls on the left and right in the fourth round?'' If I answer ``not shown,'' that's acceptable per the instructions? The instructions say to answer using only the supplied frames; if I can't localize evidence, use an empty list. It doesn't forbid answering ``not shown.''

However, these benchmark questions usually have a correct answer derivable from the frames. Let me think if there's a known streamer ``split or steal'' stream: I recall streamer played ``split or steal'' with two girls? Actually, I recall a stream where streamer played ``split or steal'' with two female streamers? Hmm. There's a famous ``split or steal'' game on streams where the streamer and a guest each choose split or steal.

Wait, maybe the ``two girls on the left and right'' refers to the two video call windows in frame 57? The top-left window shows a person in a black hoodie holding a phone (looks male). The bottom-right shows a person in a white shirt (male). Not girls.

Alternatively, frames 40-44 Omegle: frame 40 left is a boy; frame 42 left is a man with a cap; frame 43 left is a blurry person; frame 44 left is a person getting a haircut. Not girls.

Thus, indeed no two girls.

Given all this, I'll answer that the sampled frames don't show a ``Split or Steal'' game, so the choices can't be determined; evidence is empty.

But hmm, the question might be from a series where the answer is ``split'' and ``steal.'' The phrasing ``the final choices made by the two girls on the left and right in the fourth round'' suggests that in the fourth round, the left girl chose one and the right girl chose the other. If the game is ``split or steal,'' the final choices are probably ``split'' (left) and ``steal'' (right). The answer might be ``The girl on the left chose to split and the girl on the right chose to steal.''

Which is more likely as the expected answer? The benchmark probably generated the question from the actual video (with audio). The frames alone might not show the choices, but the question expects knowledge from the video. But we only have frames. The instruction: ``answer using only the sampled video frames.'' If the frames don't contain it, we should say we can't determine.

Given the risk, I'll answer: The frames don't show a ``Split or Steal'' game; the only voting polls shown are about who advances to the final round (frames at 75020.5 and 77249), so the girls' choices can't be determined. Evidence: maybe cite those poll frames? But they don't support the answer about the girls' choices. The instruction: ``the evidence field must contain at most 8 absolute video time intervals supporting the answer.'' If the answer is ``not shown,'' the evidence list should be empty.

I'll go with: answer: ``The sampled frames do not show a 'Split or Steal' game with two girls, so their fourth-round choices cannot be determined.'' Evidence: [].

Hmm, but maybe the grader expects a specific answer like ``left: split, right: steal.'' Let me weigh: the question is very specific: ``the two girls on the left and right in the fourth round.'' If the video indeed contains such a game (maybe between frames 73535-77249 where the streamer is with participants), the frames at 75020.5 and 77249 show polls that might be the fourth round's vote: ``who moves on to the final round?'' That's the final round, not the fourth.

Actually, the game might be: each round, the two girls choose split or steal; the chat votes who moves on. The fourth round's final choices might be shown in the poll? No.

I think the honest answer is that it's not determinable from the frames.

But wait, the question might be a ``trick'' question testing hallucination: the correct answer is that no such game appears, and the expected answer string might be ``The video does not show a Split or Steal game.'' The evidence is empty.

I'll commit to that.

Let me craft the JSON:

\{``answer'': ``The sampled frames do not show a 'Split or Steal' game with two girls, so their fourth-round choices cannot be determined from the provided frames.'', ``evidence'': []\}

That's concise enough.
\end{chatbox}

\noindent\begin{minipage}{\linewidth}
\begin{chatbox}{Example 2: Qwen3.8-27B prediction and score}
\textbf{Final answer:} The sampled frames do not show a 'Split or Steal' game with two girls, so their fourth-round choices cannot be determined from the provided frames.\\
\textbf{Rubric score:} 0/5. Neither required left/right choice is provided.
\end{chatbox}

\begin{chatbox}{Example 2: Codex (GPT-6.1-Sol) tool-use trajectory}
\textbf{Search.} Codex sampled the March 19 recording every 40 seconds around a candidate game segment.\\
\textbf{Closer inspection.} It sampled the candidate round every three seconds, then cropped the card area and reviewed five seconds of the reveal at five frames per second.

\par\smallskip
\textbf{Final answer:} \emph{The girl on the left chose Split, and the girl on the right chose Steal.}\\
\textbf{Cited evidence:} The card reveal at 153510--153512.\\
\textbf{Rubric score:} 5/5. Both choices and their left/right assignments match.\\
\textbf{Evidence check:} The cited card reveal falls within the annotated interval.
\end{chatbox}
\end{minipage}

\clearpage
\begin{outerbox}[title={Example 3: the first public transport in Hong Kong}]
\suppanchor{app:analysis:example-3}{Example 3: the first public transport in Hong Kong}
\begin{chatbox}{Question}
What was the first type of public transportation that the streamer took during his trip to China in Hong Kong?
\end{chatbox}

\begin{chatbox}{Reference answer and frozen rubric}
\textbf{Answer:} Bus\\
\textbf{Annotated evidence:} [283337, 284568] seconds.

\smallskip
\textbf{Accepted forms:} Bus, a bus, public bus, or bus ride. Normalize the final requested vehicle; timestamps are not answer values.\\
\textbf{5:} The normalized answer is bus.\\
\textbf{0:} Any other vehicle, missing answer, or refusal. Tram, ferry, metro/train, taxi, and private car are not accepted.\\
\textbf{Scoring:} This single-choice item uses the exact-match 5-or-0 rule; scores 1--4 are not assigned.
\end{chatbox}

\end{outerbox}
\begin{chatbox}[breakable, title after break={Example 3: Qwen3.8-27B thinking (continued)}, before skip=6pt, after skip=6pt, before upper={\setlength{\parskip}{3pt}}]{Example 3: Qwen3.8-27B complete thinking}
The user is asking about the first type of public transportation the streamer used in Hong Kong during his China trip.

Let me look at the frames to find the Hong Kong segment. The trip to China seems to start around frame 137 (180495.50 seconds), where he's in a clothing store, then the Forbidden City (frames 139-140), then a car (frames 142-143), walking in an old town (144), a rickshaw (146), a train station (149), a train (150), an escalator (151), stairs (152), a mountain (153), then a martial arts temple (154-166), then a black screen, then a car (182), a subway station with a hospital sign (183), a subway/train car (184), a crowd (185), a restaurant (186), a bus (187), a car (188), a street (189), a cosplayer (190), a martial arts demonstration (191), a car (192), a park with a tennis racket (193), a car (194), an elevator (195), a car (196), a ferry terminal (197), a river at night (198), then a red jersey ``streamer 7'' (199), a park (200), a white building (201), walking (202), a soccer field (203), a car (204-206), a bus with people hanging out the window (207), inside a bus (208), a car (209-210), a crowd (211), a river (212), a boat (213), a crowd (214), a store (215), a park (216), a car (217), a panda (218), a man (219), a stage (220), a car at night (221), a car (222), a black screen, then Mongolia (232-239), then back to the streamer.

The question is: ``During his China trip in Hong Kong, what was the first type of public transportation the streamer used?''

Hong Kong segment: frames 199-222 seem to be Hong Kong (the ``streamer 7'' red jersey, the skyline, the bus with open windows typical of Hong Kong, the Star Ferry, the Victoria Harbour skyline). Actually, the ferry terminal in frame 197 (259973.00) and the river at night in frame 198 (261459.00) might be Chongqing or somewhere else. Let me think.

The trip: Beijing (Forbidden City), then Shaolin Temple (Henan), then Chongqing? Then Hong Kong? Then Macau? Then Mongolia.

Frame 183: subway station with a hospital ad (Chongqing?). Frame 184: inside a subway car. Frame 187: a bus with people taking photos. Frame 197: a ferry terminal with a sign. Frame 198: a river with a skyline at night—could be Chongqing (the Yangtze River) or Hong Kong.

Frame 199: ``LOG-ON'' sign, red jersey streamer 7—this looks like a shopping mall area. Frame 200: a park with mountains in the background. Frame 201: a white building with a blue roof. Frame 203: a soccer field with mountains. Frame 207: a bus with the streamer hanging out the window, a Starbucks, a street with trams? Actually, frame 207 shows a yellow bus and a street with tram tracks in the middle—Hong Kong has trams. Frame 208: inside a bus with a yellow interior—Hong Kong buses are often yellow? Actually, Hong Kong minibuses are yellow. Frame 213: a boat with a skyline—Victoria Harbour. Frame 220: a stage with palm trees—could be a show.

The question asks about ``the first type of public transportation'' in Hong Kong. Looking at the sequence: after the ferry terminal (197) and the river (198), frame 199 shows him walking near a ``LOG-ON'' store. Then frame 207 shows a bus (a yellow open-sided bus?) with him hanging out the window. Frame 208 shows him inside a bus holding a phone.

But wait, before that, frame 197 shows a ferry terminal (a pier with a staircase and a sign). He might have taken a ferry. The frame shows him walking through the terminal. Frame 198 shows him on a boat deck at night looking at the skyline—this is a ferry/boat ride.

Hmm, but which city is that? The skyline with the tall red-lit building could be Chongqing (the Yangtze River) or Hong Kong (Victoria Harbour). Frame 213 shows a boat with a skyline that looks like Hong Kong (the IFC tower). Frame 212 shows a river with modern buildings—could be Hong Kong.

Let me try to order it: frames 182-198: car, subway, bus, ferry terminal, river at night. Then frames 199-222: Hong Kong segment with the red jersey.

Actually, the red jersey ``streamer 7'' appears from frame 199 onward. The question says ``during his China trip in Hong Kong.'' So the Hong Kong segment probably starts around frame 199. The first public transportation in Hong Kong: frame 207 shows a bus (a yellow double-decker? or a tram?). Let me look at frame 207: on the right there's a yellow vehicle with open windows, and the streamer is leaning out. It looks like a bus. There's also a street with tram tracks in the middle (a single line?). Actually, Hong Kong's Central has tram tracks. The vehicle on the right is yellow with a black top—Hong Kong's double-decker buses are often red or blue, but minibuses are yellow. The vehicle looks like a bus with passengers leaning out the window.

Frame 208: inside a bus with a yellow ceiling and poles, the streamer is standing holding a phone, and there's a person in a white shirt in the background. This is a bus.

But before frame 207, frames 200-206 show him walking in a park, on a soccer field, and in a car. So the first public transportation in Hong Kong would be the bus (frames 207-208).

But wait, frame 199 shows him pointing near a ``LOG-ON'' store—that's walking. Frame 207 is the bus.

Alternatively, the ferry in frames 197-198 might be part of Hong Kong (the Star Ferry). The skyline in frame 198 with the red-lit tower... Hong Kong's skyline at night includes the IFC and Central. The red-lit building on the left might be a specific tower. Hmm.

Let me think about the actual stream: streamer's China trip (April 2025?). He visited Beijing, Shaolin Temple, Chongqing, Hong Kong, Macau, Mongolia. In Hong Kong he took a bus, the Star Ferry, etc.

The question: ``the first type of public transportation the streamer used in Hong Kong.'' Based on the frames, the Hong Kong segment starts around frame 199 (262944.50). The first public transportation shown there: frame 207 (283742.00) shows him on a bus hanging out the window; frame 208 (284485.50) shows him inside a bus. So the answer: a bus.

But wait, frames 197-198 (ferry terminal and boat) might also be Hong Kong. The sign in frame 197 has Chinese characters and a staircase; it looks like a ferry pier. Frame 198 shows him leaning on a railing on a boat deck at night with a city skyline across the river. If that's Hong Kong, then the first public transportation would be a ferry/boat.

Hmm. Let me consider the geography: after Chongqing (the hotpot restaurant in frame 186? the street in frame 189?), the trip might have gone to Hong Kong. The red jersey appears in frame 199. The skyline in frame 198: a tall building with horizontal red lighting on the left—does that resemble Chongqing's Raffles City? Actually, Chongqing's skyline has the ``vertical'' Raffles City. The image shows two tall towers with red horizontal stripes—could be Hong Kong's ``Two IFC''? No.

Let me think about the actual vlog: I recall streamer's China trip included Hong Kong where he took a yellow minibus? Hmm.

Let me look at frame 207 more carefully: on the right there's a yellow bus with a black stripe, and the streamer is leaning out the open window. On the street there's a white box truck, a Starbucks, and people taking photos. There's a single rail line in the middle of the road—Hong Kong's tram tracks? Actually, the line in the middle might be a tram track. The yellow vehicle might be a tram? Hong Kong's trams are double-decker with open sides, often yellow? Historically trams were yellow? Modern trams are red/blue. The vehicle in the frame has a yellow body with a black roof and open windows where passengers lean out. It might be a ``double-decker bus'' tour? Hmm.

Frame 208: interior with a yellow ceiling, metal poles, rows of seats—looks like a bus or tram. The streamer is standing holding a phone, wearing the red jersey.

Given the question expects a specific answer like ``bus,'' the evidence would be frames 207-208.

But the question says ``the first type of public transportation... in Hong Kong.'' The sequence within Hong Kong: walking (199-202), soccer (203), car (204-206), bus (207-208), car (209-210), crowd (211), river (212), boat (213)... So the first public transportation is the bus.

But wait, is the segment from frame 199 onward Hong Kong? The ``streamer 7'' red jersey with a ``FIFA'' patch? Frame 200 shows a patch ``UNITED ... 2025''? Hard to read. Frame 205 shows a crest with a lion? The crest looks like a circular logo with a lion—could be a ``Hong Kong'' team? Hmm.

Alternatively, frames 199-213 might be Macau? The skyline in frame 212 with modern buildings across the water could be Hong Kong viewed from Macau? Frame 213 shows a boat with a skyline including a tall tower with a crown-like top? That's the IFC? Hmm.

Actually, frame 213's skyline: a cluster of tall buildings with a distinctive tower on the left (a tall rectangular one)—looks like Hong Kong's Victoria Harbour skyline (the IFC). The boat might be a ferry.

The question is probably from a quiz about the stream: ``What was the first type of public transportation the streamer used in Hong Kong?'' The answer is probably ``bus'' (the yellow bus/minibus). The evidence frames are 207-208.

But let me double-check earlier frames for Hong Kong: frames 183-184 show a subway (Chongqing's subway is famous). Frame 187 shows a bus with fans outside (Chongqing?). Frame 197 shows a ferry terminal (Chongqing's Yangtze River ferry?). Frame 198 shows a night river cruise (Chongqing). So Hong Kong starts at 199.

In frame 199, he's wearing the red jersey and pointing. Then frames 200-202 walking. 203 soccer field. 204-206 car. 207 bus. So the first public transportation in Hong Kong = bus.

Hmm, but frame 199's ``LOG-ON'' sign is a fashion store chain in Asia; not relevant.

Alternatively, the first public transportation in Hong Kong might be a ``tram'' (the double-decker tram). The vehicle in frame 207: let me look again: yellow with a black roof, open windows, passengers leaning out. Hong Kong's trams are enclosed double-deckers with windows, not open. The vehicle looks like a bus with open windows where the streamer leans out. Frame 208's interior shows a yellow ceiling and poles, with rows of seats on both sides—typical of a bus.

I think the answer is: a bus (a yellow bus/minibus).

But wait, the question might be simpler: the first public transportation in Hong Kong shown is the bus in frame 207 (283742.00) and 208 (284485.50). So the answer: ``a bus.''

Let me double-check frame 207: on the right, a yellow vehicle with ``windows'' and people inside; the streamer is leaning out the window holding the door frame. Yes, it's a bus.

Therefore the answer: a bus.

Evidence: [[283742.0, 284485.5]] maybe also include 284485.5-286713.5. Provide at most 8 intervals; one is enough: [283742.00, 285971.0]? Frame 208 is 284485.5. Let me give [283742.0, 285971.0] covering frames 207-208? Frame 209 is 286713.5 (car). So the interval [283742.0, 286713.5] covers the bus interior and the start of the car. Better: [283742.0, 285971.0].

Hmm, but the boundary must be a decimal; no problem.

Actually, let me double-check whether the ferry in frames 197-198 is Hong Kong. If the Hong Kong segment started at 197, then the first public transportation would be a ferry/boat. The sign in frame 197: a blue sign with Chinese characters and a staircase; the place looks like a ferry pier. Frame 198: a night boat with a skyline. The skyline in 198: on the left a tall building with red horizontal lighting, other lit buildings, a bridge in the distance. Hong Kong's Victoria Harbour at night: the skyline includes many lit buildings; the red-lit building might be ``The Center''? Hmm.

But frame 198's river is wide with a bridge in the middle distance—Victoria Harbour has bridges (the Tsing Ma is far). The scene might be a river cruise in Chongqing (the Yangtze). Chongqing is known for night river cruises. And frame 186's restaurant with a menu board ``Shi Hui Fan Dian'' (a cheap restaurant) is probably Chongqing. Frame 189's street with a wooden pavilion is probably Chongqing. So 182-198 = Chongqing. Then 199 onward = Hong Kong.

In Hong Kong, the first public transportation shown: the bus in 207.

But wait, frame 199 shows him outside a mall; 200 a park; 201 a white building (maybe a ``ice'' theme?); 202 walking with a crowd; 203 a soccer field; 204-206 a car; 207 a bus. Yes.

So the answer: a bus.

Let me also consider that the vehicle in frame 207 might be a ``tram.'' The street shows a single line in the middle—could be a tram track. But the vehicle is on the right side of the road, not on the tracks. The line in the middle might just be a road marking. The vehicle is yellow with a black top; Hong Kong's double-decker buses are often yellow? KMB buses are red/blue; Citybus is blue; New World First is... yellow is the color of minibuses. The vehicle looks like a single-decker with open windows? Hard to tell.

The question probably expects ``bus.'' I'll answer bus.

Evidence intervals: [283742.0, 284485.5] (frame 207) and [284485.5, 286713.5] (frame 208). Combine into [283742.0, 286713.5]? But 286713.5 is frame 209 (car). The end boundary can be 286713.5 as the end of the bus segment. Let me use [283742.0, 286713.5].

Hmm, but the interval should support the answer; the bus appears in frames 207 and 208 (283742.0 and 284485.5). So the interval [283742.0, 285971.0] is fine.

I'll provide the answer ``a bus'' with evidence [[283742.0, 285971.0]].

Wait, but the question says ``the first type of public transportation... in Hong Kong.'' If the ferry in 197-198 was Hong Kong, the answer would be ``ferry/boat.'' Let me examine frame 198's skyline more: on the left a tall tower with red horizontal stripes (two of them). In Hong Kong, the ``Two IFC'' is a single tower; the ``Central Plaza'' is... The red-lit building might be ``The Ritz-Carlton''? Hmm.

Actually, I recall from the stream: in Chongqing he took a night river cruise? I'm not sure. Let me think about the actual itinerary from the video: after Shaolin (154-166), black screen, then 182 car with him in a gray jacket, 183 subway station, 184 subway, 185 crowd, 186 street restaurant, 187 bus, 188 car, 189 street, 190 cosplayer, 191 martial arts, 192 car, 193 park, 194 car, 195 elevator, 196 car, 197 ferry terminal, 198 river at night. This is probably Chongqing (the subway, the hotpot, the night cruise). Then 199 onward is Hong Kong with the red jersey.

In Hong Kong: 199 pointing, 200 walking, 201 white building, 202 walking, 203 soccer, 204-206 car, 207 bus, 208 bus interior, 209-210 car, 211 crowd with a man in a red floral jacket, 212 river with a family, 213 boat, 214 crowd, 215 store, 216 park, 217 car, 218 panda, 219 man, 220 stage, 221 car at night, 222 car.

So the first public transportation in Hong Kong = bus (207-208). Good.

Thus the final answer: a bus.

Evidence: [[283742.0, 285971.0]].

Maybe also include frame 208's timestamp 284485.5 within the same interval.

I'll output the JSON.
\end{chatbox}

\noindent\begin{minipage}{\linewidth}
\begin{chatbox}{Example 3: Qwen3.8-27B prediction and score}
\textbf{Final answer:} A bus (he is seen riding a yellow bus, leaning out the window and then standing inside it)\\
\textbf{Rubric score:} 5/5. Bus is the required normalized vehicle; the additional description is compatible.
\end{chatbox}

\begin{chatbox}{Example 3: Codex (GPT-6.1-Sol) tool-use trajectory}
\textbf{Search.} Codex sampled the Hong Kong recording every 15 minutes to find the transit scenes.\\
\textbf{Closer inspection.} It sampled the candidate ride every two minutes and extracted a full-resolution frame to inspect the vehicle.

\par\smallskip
\textbf{Final answer:} \emph{He took a double-decker tram.}\\
\textbf{Cited evidence:} Frames at 283311, 283431, and 284031.\\
\textbf{Rubric score:} 0/5. The required vehicle is a bus.\\
\textbf{Evidence check:} Two of the three cited timestamps fall within the annotated interval.
\end{chatbox}
\end{minipage}

\clearpage
\begin{outerbox}[title={Example 4: aggregating the boat-trip locations}]
\suppanchor{app:analysis:example-4}{Example 4: aggregating the boat-trip locations}
\begin{chatbox}{Question}
In which places did the streamer take a boat during his entire trip to China?
\end{chatbox}

\begin{chatbox}{Reference answer and frozen rubric}
\textbf{Answer:} Hong Kong and Chongqing\\
\textbf{Annotated evidence:} [293942, 296981]; [258380, 260187] seconds.

\smallskip
\textbf{Accepted forms:} Hong Kong and Chongqing in either order; Hong Kong SAR is an alias. Free-form answers receive semantic grading.\\
\textbf{5 / 4:} Both places with no incompatible addition / correct core with only a minor wording or compatible specificity issue.\\
\textbf{3:} Substantial correct place coverage, but an incomplete or imperfect set.\\
\textbf{2:} A correct place mixed with an important error or contradiction.\\
\textbf{1 / 0:} Only boating-related overlap without a target place / no requested fact, unrelated or incoherent content, or refusal.\\
\textbf{Rejections:} Other cities do not substitute for either target; generic river or waterfront descriptions are insufficient.
\end{chatbox}

\end{outerbox}
\begin{chatbox}[breakable, title after break={Example 4: Qwen3.8-27B thinking (continued)}, before skip=6pt, after skip=6pt, before upper={\setlength{\parskip}{3pt}}]{Example 4: Qwen3.8-27B complete thinking}
\textit{This rerun allows 16384 generation tokens after an initial 6000-token truncation.}

The user is asking about the places where the streamer took a boat during his China trip. Let me look through the frames and find boat scenes.

Frame 197 (259973.0 seconds): The streamer is in a ferry terminal with stairs and railings, looks like a pier/wharf. There's a sign that appears to be in Chinese. He's walking through what looks like a ferry terminal.

Frame 198 (261459.0 seconds): The streamer is on a boat at night, looking at the river with the city skyline (looks like Chongqing, with the red-lit building). He's leaning on the railing of a boat with blue LED lights. So he took a boat on the river (Jialing River/Yangtze River in Chongqing).

Frame 212 (294141.0 seconds): The streamer is on a boat with people, with the city skyline and water in the background. Looks like a river cruise boat in Chongqing.

Frame 213 (296369.5 seconds): The streamer is seated on an open-air boat deck, reading paper, with the city skyline and river at night in the background. This is also a boat ride.

Frame 214 (298598.0 seconds): The streamer is in the foreground with a crowd, probably getting off the boat.

Let me check other frames for boats. Frame 197 shows a ferry terminal (stairs, railings, ``Ferry'' sign?). Actually, frame 197 shows a terminal with stairs and a sign, probably a ferry boarding area.

So the boat scenes: night river cruise in Chongqing (frames 198, 212, 213). Are there other boat rides? Let me scan all frames.

Frames 0-32: Elden Ring gameplay. 33-37: Connection error. 38-46: Streamer in room, OmeTV. 47-50: Games. 51-59: OmeTV. 60-61: Black screen. 62: CBS Sports studio. 63-81: Football training/match (Brighton). 82-83: Connection error. 84-87: Quiz, OmeTV. 88-99: Roblox. 100-106: Discord, browsing, unboxing. 107-111: Overwatch 2. 112-119: Green screen, OmeTV. 120-136: Connection error. 137-153: China trip begins - clothing store, Forbidden City (Beijing), red floral suit, rickshaw, subway. 154-166: Shaolin temple training. 167-181: Connection error. 182-196: Chongqing? Subway, street, bus, elevator. 197: Ferry terminal. 198: Night boat on river. 199-211: Hong Kong? Red jersey ``streamer 7'', street, bus, car. 212-213: Boat with city skyline (Chongqing river cruise). 214-222: More China scenes, panda plushie, performance. 223-231: Black screen. 232-239: Mongolia? (khan, horse, wrestling). 240-249: Back to stream, FIFA, GTA.

So the boat scenes: frame 198 (261459.0 seconds) night river boat; frames 212 (294141.0 seconds) and 213 (296369.5 seconds) open-air boat with skyline. Frame 197 (259973.0 seconds) shows the ferry terminal boarding area.

The question is: ``During the entire China trip, where did the streamer take a boat?'' Probably the answer is: on the river in Chongqing (night river cruise) — the boat was taken in Chongqing. But maybe also in Hong Kong? Frame 198 shows a river with red-lit skyscrapers, which is Chongqing (the red-lit building is the Chongqing skyline, e.g., Raffles City? Actually, the red-lit twin towers look like Chongqing's ``Raffles City''? Hmm). Frame 212 shows a modern city skyline with water — could be Chongqing. Frame 213 shows a night skyline with lights — Chongqing.

Wait, the sequence: frames 199-211 show a red jersey with ``streamer 7'' and a ``FIFA'' patch, street scenes with a Starbucks, a yellow open-top bus (frame 207), which looks like Hong Kong (the tram-like yellow bus? Actually, frame 207 shows a yellow double-decker-like bus with people leaning out the window, typical of Hong Kong open-top sightseeing buses). Frame 203 shows an outdoor court with mountains in the background (Hong Kong). Frame 200 shows a large white tent structure (could be a Hong Kong venue). So frames 199-211 are Hong Kong.

Then frames 212-213 show a boat with skyline — could be a Chongqing river cruise after returning? Hmm, but the order: frame 211 (292655.5 seconds) shows the streamer in a red jersey in a crowd with a man in a red floral jacket. Frame 212 (294141.0 seconds) shows the streamer in a red jersey on a boat with the skyline. Frame 213 (296369.5 seconds) shows the streamer in a red jersey seated on a boat at night. So the boat ride is while wearing the red jersey, which was worn in the Hong Kong segment? But the skyline in 212-213 looks like a river city with tall buildings — could be Chongqing. Hmm.

Let me think about the actual video: This is streamer's China trip (April 2024). He visited Beijing (Forbidden City), Shaolin temple, Chongqing, Hong Kong, Mongolia, etc. In Chongqing, he took a river cruise on the Yangtze/Jialing River at night. In the video, the night boat scene with the red-lit skyline is Chongqing.

But wait, frame 198 (261459.0 seconds) shows the streamer in a gray robe (Shaolin-style jacket) on a boat at night with the red-lit skyline. That's after the Shaolin temple segment (frames 154-166) and the Chongqing street scenes (frames 182-196). So frame 198 is a Chongqing river cruise.

Then frames 199-211: red jersey — Hong Kong. Frame 212: red jersey on a boat with skyline — hmm, but that might also be Chongqing? The order in the video might be non-linear (edited). Frame 212 shows a daytime/overcast river with modern high-rises — could be a Hong Kong harbor? The skyline in 212 shows tall glass buildings along the water, could be Chongqing's riverside. Frame 213 shows a night skyline with many lights — Chongqing.

Actually, wait: In the real video, streamer's Chongqing segment included a night boat cruise on the river. The Hong Kong segment included street food, etc. The boat scenes with the red jersey might be a Chongqing river cruise that happened after Hong Kong? The trip order was: Beijing \ensuremath{\rightarrow} Chongqing \ensuremath{\rightarrow} Hong Kong \ensuremath{\rightarrow} Mongolia? Hmm.

Let me try to recall the actual vlog: streamer's China trip in April 2024: He started in Beijing (Forbidden City, red suit), then Shaolin temple (Henan), then Chongqing (river cruise, hotpot), then Hong Kong (street, bus), then Mongolia (khan, horse). Actually, I think the order was Beijing, then Shaolin, then Chongqing, then Hong Kong, then Mongolia. The boat cruise in Chongqing was at night on the river.

Given the frames: Frame 197 (259973.0 seconds) shows a ferry terminal with stairs — that's probably the Chongqing ferry terminal. Frame 198 shows the night boat. Frames 212-213 show the boat again (maybe the same cruise, different times). The red jersey in 212-213 suggests it's after the Hong Kong segment? Hmm, but the red jersey ``streamer 7'' with the FIFA patch was worn in the Hong Kong frames 199-211. In 212-213 he's still wearing the red jersey. So the boat ride in 212-213 might be in Chongqing after returning from Hong Kong? Or the video editing is mixed.

Alternatively, the boat in 212-213 might be in Hong Kong? The skyline in 212: tall modern buildings, overcast sky, water — could be Hong Kong's Victoria Harbour? But the buildings don't look like HK's iconic skyline (no IFC or Central towers visible). Frame 213's night skyline with lights along the river looks like Chongqing.

Hmm, but the question is: ``During the entire China trip, where did the streamer take a boat?'' Probably the expected answer: ``On the river in Chongqing (a night river cruise).'' The evidence frames: 197 (terminal), 198 (boat), 212, 213 (boat). Maybe also 214 (crowd after getting off the boat).

But maybe the question expects two places: Chongqing (river cruise) and Hong Kong? Let me check if there's a boat in Hong Kong: Frame 207 shows a yellow bus, not a boat. Frame 208 shows the interior of a bus. No boat in HK frames.

Frame 220 (313453.5 seconds) shows a performance stage with palm tree background — that's a concert, not a boat.

So the boat is only in Chongqing. But wait, frame 198's skyline: two red-lit towers — that's Chongqing's ``Raffles City''? Actually, Chongqing's red-lit building might be the ``Chongqing Raffles City'' complex? Hmm, the red-lit twin towers in frame 198 look like the ``Chongqing'' skyline at night. Also frame 240 (359506.0 seconds) shows the same red-lit twin towers at night with the streamer posing with a woman in red — that's Chongqing.

So the answer: He took a boat on the river in Chongqing (a night cruise on the Yangtze/Jialing River). The question says ``where did the streamer take a boat during the entire China trip?'' Probably the answer: ``In Chongqing, on the river (a night boat cruise).''

But maybe they want both: the ferry terminal and the river cruise are the same place. So the answer: Chongqing (river cruise).

Let me double-check frame 197: It shows a terminal with white beams, stairs, railings, and a sign with Chinese characters ``Ferry...''? The sign reads ``Ferry...''? Hard to read. It looks like a ferry boarding area. So that's the Chongqing ferry terminal.

Therefore evidence intervals: [259973.0, 261459.0]? Actually, the intervals should be ranges covering the boat scenes: 197: 259973.0; 198: 261459.0; 212: 294141.0; 213: 296369.5. I can give intervals like [259973.0, 261459.0], [294141.0, 296369.5]. But each interval must have end > start. I can use [259973.0, 262944.5]? Hmm, better to give small ranges around the frames: e.g., [259973.0, 261459.0] covers the terminal and boat. But the boat scene is at 261459.0; the next frame 199 at 262944.5 is a street. So the boat scene is roughly between 261459 and 262944.5. Similarly, 212-213 are between 294141 and 296369.5, and 214 at 298598 is a crowd.

Let me construct the evidence: [[259973.0, 262944.5], [294141.0, 298598.0]]. This covers the terminal + boat and the boat scenes.

But the question might expect the answer ``Chongqing'' as the place. Let me phrase it: ``On the river in Chongqing (a night river cruise from the ferry terminal).''

Alternatively, the question might be interpreted as: ``Where did he take a boat?'' Answer: ``In Chongqing, on the river (Jialing/Yangtze River cruise).''

I'll answer: He took a boat only in Chongqing, on a night river cruise (boarding at the ferry terminal).

Evidence: [[259973.0, 262944.5], [294141.0, 298598.0]].

Wait, but frame 212 shows an overcast daytime scene; frame 213 shows night. Both are on a boat. No problem.

Also, is there a boat in frame 220? No, that's a stage.

So the final answer: Chongqing (river cruise).

Let me double-check the red-lit twin towers in frame 198: That's probably the ``Chongqing'' skyline near the Raffles City complex? Actually, the red-lit towers look like the ``Chongqing'' skyline. Also frame 240 shows the same towers with the streamer and a woman — caption ``a viewer has gifted 10 Tier 1 memberships'' — that's a stream overlay, and the background is a still image of Chongqing.

OK.

So the JSON: \{``answer'': ``Only in Chongqing – he boarded a ferry/river cruise boat on the city's river (night cruise).'', ``evidence'': [[259973.0, 262944.5], [294141.0, 298598.0]]\}

But the question says ``where did the streamer take a boat during the entire China trip?'' Maybe they expect the answer ``Chongqing'' as the single place. I'll include the detail.

Make sure the boundaries are decimals: 259973.0 etc. are fine.

Let me provide at most 8 intervals; I'll provide 2.

Final answer.
\end{chatbox}

\noindent\begin{minipage}{\linewidth}
\begin{chatbox}{Example 4: Qwen3.8-27B prediction and score}
\textbf{Final answer:} The streamer only took a boat in Chongqing, where he boarded a ferry/river cruise boat on the city's river (seen at the ferry terminal and on the open-deck night cruise with the lit skyline).\\
\textbf{Rubric score:} 2/5. Chongqing is correct, but 'only' explicitly excludes the required Hong Kong visit: a correct place mixed with an important contradiction.
\end{chatbox}

\begin{chatbox}{Example 4: Codex (GPT-6.1-Sol) tool-use trajectory}
\textbf{Search.} Codex scanned ten recordings at 15-minute intervals for possible boat rides.\\
\textbf{Closer inspection.} It sampled the candidate Chongqing cruise every 200 seconds and the candidate Hong Kong cruise every 300 seconds.

\par\smallskip
\textbf{Final answer:} \emph{He took boats in Chongqing and Hong Kong.}\\
\textbf{Cited evidence:} Chongqing at 260266, 261066, and 261866. Hong Kong at 294051, 294951, and 296751.\\
\textbf{Rubric score:} 5/5. The answer names both required locations.\\
\textbf{Evidence check:} The Hong Kong timestamps fall within the annotated interval for that ride. The Chongqing timestamps fall after the annotated interval for that ride.
\end{chatbox}
\end{minipage}

\clearpage
\section{Limitations}
\label{app:limitations}

\paragraph{(i) Single subject per scale.}
Each scale in our benchmark tracks one primary subject: one game character on the Day scale, one camera wearer on the Week scale, and one live streamer on the Month scale. This depth-over-breadth choice provides substantial temporal depth and ensures that solving the benchmark requires reasoning about a single agent's evolving state, rather than aggregating shallow clips across many unrelated subjects (which is already covered by existing multi-video benchmarks~\cite{peng2025mvu-eval,zhu2025cvbench}). The three scales also draw from disjoint subjects in disjoint domains (gamer, wearer, streamer), so a model cannot solve all three by overfitting to one persona. We acknowledge that any per-scale claim still rests on a single subject's stream, and whether the failure modes we observe generalize to additional subjects at the same scale is an empirical question that requires further longitudinal recordings. Recruiting and clue-grounding multi-week streams from $N>1$ subjects scales annotator-months roughly linearly in $N$, which is difficult to justify before the community has begun saturating the benchmark in its current form. Extending the per-scale subject pool is therefore left to future versions of the benchmark.

\paragraph{(ii) Month-scale upper bound on $T_{span}$.}
We cap the physical temporal span at the month scale. While this is sufficient to expose the central failure modes documented in this paper, genuinely lifelong understanding will ultimately require maintaining coherent state across much longer horizons: identity drift, recurring routines, and major life events that play out over years rather than weeks. Extending MM-Lifelong toward such horizons is an open direction left to future versions of the benchmark.

\paragraph{(iii) The benchmark does not explicitly probe unobserved-interval reasoning.}
A core property of the lifelong setting is that events occurring during unobserved gaps (sleep cycles, off-camera periods) still influence the observed timeline. Our current QA pairs test whether a model can connect observed events across such gaps via multi-hop reasoning, but the benchmark does not contain questions or annotations that directly probe the content of an unobserved interval itself. As a result, a model's ability to infer what happened during a gap is only indirectly assessed through whether its answers are consistent with the observed evidence on either side. Building probes that directly test gap-content inference requires both new task formulations (e.g., counterfactual queries about what happened during a gap) and ways of validating ground truth without access to the unrecorded footage. We see this as the most natural direction for follow-up work building on the released benchmark; finer-grained temporal analysis such as action segmentation~\cite{huang2020improving} and egocentric attention modeling~\cite{huang2018predicting} may also be useful for capturing within-segment dynamics that complement inter-gap inference.

\section{Ethics Statement}
\label{app:ethics}

\paragraph{Data subjects and consent.}
MM-Lifelong contains three data subjects: a synthetic game protagonist (no privacy concern), a camera wearer whose footage is sourced from EgoLife~\cite{yang2025egolife} under their original consent protocol, and a public live-streamer whose broadcasts are publicly accessible on the originating platform. As detailed in Section~\ref{subsec:comparison}, we do \emph{not} redistribute any raw video file; users obtain footage from the original platform under that platform's terms of use. This design keeps PII-bearing video content under the control of its origin platform, eliminating the need for additional face or voice processing on our side.

\paragraph{Third-party individuals and anonymization.}
Individuals other than the primary subject may incidentally appear in the videos. Because we do not redistribute visual content, we instead anonymize them at the \emph{annotation} layer: all named individuals in our QA pairs and metadata are replaced by role tags (e.g., ``[The Streamer]'', ``[Friend A]'', ``[Passerby]'') before release. No real names, addresses, phone numbers, or identification documents appear in any released annotation.

\paragraph{Annotator compensation and IRB.}
Eight annotators, recruited as part-time research assistants from the authors' institution, were compensated at approximately 50/hour, well above the local minimum wage. Per institutional policy, this work does not require IRB review because annotators interact only with publicly available video content and do not involve any human-subject experimentation.

\paragraph{Potential misuse and safeguards.}
Long-horizon multimodal memory datasets carry inherent risks of enabling surveillance, stalking, or cross-session re-identification of named individuals. We adopt the following safeguards to limit such misuse:
(i) we do \emph{not} redistribute raw video files, eliminating direct access to visual biometric signals through our release;
(ii) all named individuals in our annotations are anonymized via role tags, as described above;
(iii) our Acceptable Use Policy, posted alongside the benchmark, explicitly prohibits downstream use for surveillance, stalking, or commercial re-identification of any individual;
(iv) we release only text-level annotations and timestamps, with no face, voice, or other biometric embeddings;
(v) we accept takedown and removal requests through HuggingFace Discussions and commit to responding within a reasonable timeframe.

\section{Compute Resources}
\label{app:compute}

Locally hosted MLLMs and agent backbones were evaluated on shared NVIDIA A100 80GB and H100 80GB GPU clusters. Closed-source systems were accessed through their official APIs. End-to-end cost varies substantially with context length, model size, serving framework, and the number of retrieval or tool calls made by an agent. We therefore report the hardware pool and evaluation settings, while avoiding a single runtime estimate that would not be comparable across system families.

\end{document}